\documentclass[11pt]{article}

\usepackage[final]{acl}

\usepackage{times}
\usepackage{latexsym}

\usepackage[T1]{fontenc}
\usepackage[utf8]{inputenc}
\usepackage{microtype}
\usepackage{inconsolata}

\usepackage{hyperref}       
\usepackage{url}            
\usepackage{booktabs}       
\usepackage{graphicx}       
\usepackage{subcaption}     
\usepackage{placeins}       
\usepackage{amsmath}        
\usepackage{amsfonts}       
\usepackage{nicefrac}       
\usepackage{xcolor}         
\usepackage{colortbl}       
\usepackage{multirow}       
\usepackage{pifont}         
\usepackage{fvextra}        

\title{With Tools or Because of Tools? Tracing Capability Gains in Multimodal Agents}

\author{First Author \\
  Affiliation / Address line 1 \\
  Affiliation / Address line 2 \\
  Affiliation / Address line 3 \\
  \texttt{email@domain} \\\And
  Second Author \\
  Affiliation / Address line 1 \\
  Affiliation / Address line 2 \\
  Affiliation / Address line 3 \\
  \texttt{email@domain} \\}

\author{
    \textbf{Jiawei Guo\textsuperscript{1,2}\footnotemark[1],}
    \textbf{Donglei Yu\textsuperscript{1,3}\footnotemark[1]\footnotemark[3],}
    \textbf{Yu Chen\textsuperscript{2},}
    \textbf{Xiang Wang\textsuperscript{2}\footnotemark[2],}
    \textbf{Shuai Li\textsuperscript{2},}
    \textbf{Xinpei Zhao\textsuperscript{2},}\\
    \textbf{Huaxing Liu\textsuperscript{2},}
    \textbf{Qinghao Wang\textsuperscript{4},}
    \textbf{Minpeng Liao\textsuperscript{3}\footnotemark[2]}
    \\
    \textbf{\textsuperscript{1}} University of Chinese Academy of Sciences
    \quad
    \textbf{\textsuperscript{2}} Amap, Alibaba Group \\
    \textbf{\textsuperscript{3}} Tongyi Lab, Alibaba Group
    \quad
    \textbf{\textsuperscript{4}} Peking University
}

\begin{document}
\maketitle

\renewcommand{\thefootnote}{\fnsymbol{footnote}}

\footnotetext[1]{\ \ Equal contribution.}
\footnotetext[2]{\ \ Corresponding author.}
\footnotetext[3]{\ \ Intern at Tongyi Lab, Alibaba Group.}

\begin{abstract}
Tool-augmented multimodal agents show strong benchmark gains, often
taken as evidence that agents have learned to use tools. We argue that
this interpretation can be premature: a tool-call trace alone does not
show whether the tool supplied answer-critical information. We study
two representative ``thinking with images'' agents, Thyme and
DeepEyesV2, across real-world understanding, OCR, chart understanding,
and mathematical reasoning. Each agent is compared with its Tool-Free
counterpart and with a Pure-Text Reasoner trained from the same source
pool without tool-calling trajectories. Tool access yields little
consistent aggregate improvement, does not reliably reduce
generated-token cost, and leaves only a small tool-only solved set:
93\% of DeepEyesV2's tool-solved problems and 96\% of Thyme's are also
solved by at least one non-tool setting. Mechanism ablations further
show that the full tool-use loop does not consistently outperform
either the tool-call format or the returned execution result alone. In the settings we study, the analyzed agents appear to learn tool-calling patterns more reliably than tool-contributed capabilities, suggesting
that evaluation should distinguish tool availability from whether tools
actually expand what agents can solve.
\end{abstract}

\section{Introduction}

A multimodal agent is asked what material a glove in a
street-food image is made of. It writes a Python snippet,
calls the code interpreter, and answers. Yet the snippet
does little: it simply reloads and displays the original image,
with no cropping, zooming, or visual processing. A pure-text
model, given no tools, answers correctly on the first try. The
trajectory contains a tool call, but the call supplies no new
visual evidence.

\begin{figure}[t]
  \centering
  \includegraphics[width=0.95\linewidth]{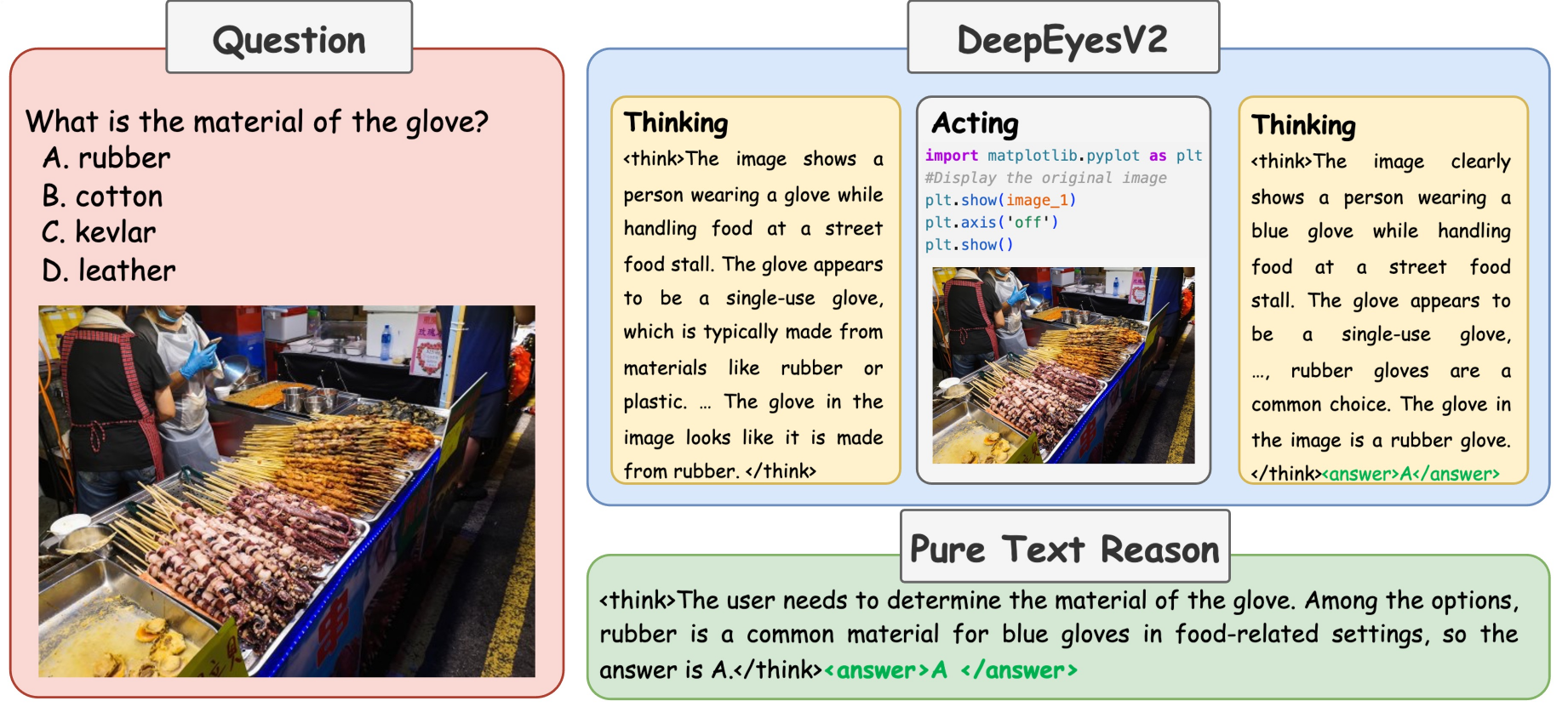}
  \caption{A motivating example illustrating that tool interaction can become procedural rather than genuinely helpful: the agent performs extra tool-mediated steps, yet this added interaction does not necessarily improve the final reasoning outcome.}
  \label{fig:intro-case}
\end{figure}


This is not an isolated curiosity. Recent multimodal systems increasingly use intermediate visual or computational operations---including OCR, code execution, region inspection, visual manipulation, or code-mediated reasoning---and their benchmark gains are often linked to these operations~\citep{wu2025qwenimagetechnicalreport,11444749,wang-etal-2025-mathcoder,wei2025deepseek,li2025tir}. Yet observing such an operation only shows that the model performed one; whether it supplied missing information, or whether the model would have failed without it, is a separate question. We therefore ask: across instances, does tool access actually expand the set of problems an agent can solve?

To examine this question, we first place tool-augmented
thinking-with-images agents in a broader benchmark comparison. 
Across real-world understanding, OCR, chart understanding,
and mathematical reasoning benchmarks, we compare several recent
tool-augmented multimodal systems with non-tool reasoning
baselines~\citep{bai2025qwen25vltechnicalreport,NEURIPS2025_2c84844a,meng2025mmeurekaexploringfrontiersmultimodal}
and a tool-free Pure-Text Reasoner trained on filtered data from the
DeepEyesV2~\citep{hong2026deepeyesv} source pool. The comparison shows that
tool-augmented agents often achieve strong results, but similar
performance can also be obtained without tool-enabled inference.

We then take a closer look at two representative thinking-with-images
agents, Thyme~\citep{zhang2026thyme} and DeepEyesV2, through solved-set, process-level, and
trajectory-ablation analyses. We choose these two systems because they
are among the strongest recent open-source agents in our benchmark
suite, expose executable tool-use interfaces, and provide sufficiently
complete model and inference resources to support fine-grained
diagnostic analysis. This choice lets us move beyond aggregate
benchmark scores and ask whether tool access changes which examples
the agents can solve. Importantly, our main attribution results do not rely solely on the Pure-Text Reasoner: they are also supported by the Tool-Free
Agent, which uses the released agent itself with tool calls suppressed
only at inference, and by format-only/result-only ablations that require
no retraining.





Three findings emerge. First, tools do not consistently improve
accuracy or reduce generated-token cost: removing tool access leaves
performance largely unchanged, while the Pure-Text Reasoner remains
competitive and sometimes stronger. Second, the tool-only solved set is
small: 93.2\% of DeepEyesV2's tool-solved problems and 96.0\% of
Thyme's are also solved by at least one non-tool setting. Third, the
agents show no stable capability-expanding tool use: DeepEyesV2 is
better preserved by the tool-call format, whereas Thyme is better
preserved by the returned result, yet neither pattern yields stable
solved-set expansion.

These findings do not imply that tools are useless. Rather, they suggest that the agents studied here learn the protocol of tool calling more reliably than the conditions under which tool outputs change the answer. This matters for evaluation: a correct answer with a tool-call trace is not, by itself, evidence that the tool mattered. Distinguishing tool availability from tool-contributed capability requires asking whether tool use changes what a model can solve.

We make three contributions. First, we reframe tool-use evaluation around
whether tool access expands an agent's solvable region, comparing each
Tool-Enabled Agent with its inference-time Tool-Free counterpart and a
separately trained Pure-Text Reasoner constructed without tool-calling
trajectories. Second, we pair sample-level decomposition of tool-call
outcomes---preserved successes, unrepaired failures, effective corrections,
and induced errors---with process-level attribution of what tool outputs
actually contribute. Third, we introduce format-only and result-only
trajectory ablations to separate dependence on the tool-calling protocol
from dependence on the returned execution result.

\section{Related Works}

\begin{table*}[!t]
  \centering
  \caption{Main perception benchmark results across real-world understanding, OCR, and chart-related tasks. The bold numbers indicate the best performance on each benchmark, and the underlined numbers indicate the second best.}
  \label{tab:main-perception}
  \small
  \setlength{\tabcolsep}{4pt}
  \resizebox{\textwidth}{!}{%
\begin{tabular}{lc | cccc | cc | ccc}
\toprule
Model & Tool & \multicolumn{4}{c|}{Real-World Understanding} & \multicolumn{2}{c|}{OCR} & \multicolumn{3}{c}{Chart} \\
\specialrule{0em}{0pt}{1pt}
\cline{3-6}
\cline{7-8}
\cline{9-11}
\specialrule{0em}{0pt}{1pt}
& & \shortstack{V* \\ Bench} & \shortstack{HRBench \\ 4K} & \shortstack{HRBench \\ 8K} & \shortstack{Tree \\ Bench} & \shortstack{OCR \\ Bench} & \shortstack{SEED \\ 2 PLUS} & \shortstack{CharXiv \\ descriptive} & \shortstack{CharXiv \\ reasoning} & \shortstack{Chart \\ QA} \\
\midrule
Qwen2.5-VL     & \ding{55} & 78.5 & 71.6 & 67.9 & 37.0 & 864 & \underline{70.4} & 72.7 & 40.2 & 86.2 \\
Pixel-Reasoner & Crop      & \underline{84.3} & 74.0 & 66.9 & 39.0 & -     & -    & -    & -    & -    \\
DeepEyes       & Crop      & \textbf{85.6} & 75.1 & \textbf{72.6} & 37.5 & -     & -    & -    & -    & -    \\
Thyme          & Code      & 82.7 & 77.0 & \underline{72.5} & 38.8 & 865 & 70.2 & 73.3 & 43.8 & \underline{87.7} \\
Thyme          & Tool-Free & 83.8 & \underline{78.5} & 72.3 & 37.5 & \underline{875} & 69.7 & 73.3 & 43.0 & 87.2 \\
DeepEyesV2     & General   & 82.2 & 76.1 & 72.3 & \underline{41.5} & 814 & 69.8 & \underline{75.7} & \underline{45.9} & \textbf{88.1} \\
DeepEyesV2     & Tool-Free & 80.6 & 74.8 & 71.1 & 33.1 & 866 & \textbf{71.2} & 77.0 & \textbf{46.2} & 83.3 \\
Pure-Text Reasoner & \ding{55} & 83.2 & \textbf{77.3} & \underline{72.5} & \textbf{42.2} & \textbf{888} & \underline{70.7} & \textbf{77.5} & \textbf{46.2} & 83.4 \\
\bottomrule
\end{tabular}%
  }
\end{table*}

\subsection{Tool-Augmented Multimodal Reasoning}

A growing line of work moves multimodal models from passive image
understanding toward active visual reasoning, often described as
``thinking with images''~\citep{openai2025o3}. Training-free systems
orchestrate vision modules through executable
programs~\citep{gupta2023visual,suris2023vipergpt} or produce visual
artifacts as intermediate reasoning steps~\citep{hu2024visual}. Recent
trained systems teach multimodal models to invoke tools directly,
covering end-to-end RL elicitation~\citep{zheng2026deepeyes},
pixel-space and zoom-and-focus operations~\citep{su2026pixel,zhang2025adaptivechainoffocusreasoningdynamic},
query-conditional tool selection~\citep{huang2025visualtoolagent},
interleaved visual editing~\citep{wu2026vtoolr}, deep multi-turn
trajectories~\citep{lai2026minio}, and refinements in trajectory
synthesis, tool-grounded reward modeling, and experience
reuse~\citep{ashraf2025matrix,ding2026arm,wang2026museagentmultimodalreasoningagent}. The two
systems we study, Thyme~\citep{zhang2026thyme} and DeepEyesV2~\citep{hong2026deepeyesv}, follow
this paradigm by using code execution for image manipulation and
numerical computation during reasoning. Prior work asks how to improve
tool use; we ask whether tool access expands what agents can solve.

\subsection{Faithfulness and Necessity of Intermediate Reasoning}

A parallel literature questions whether tool-mediated reasoning
steps are faithful to, or necessary for, the final answer~\citep{turpin2023language}.
MME-CoT~\citep{jiang2025mmecot} shows that high final-answer
accuracy can coexist with logically broken intermediate steps, and
Faithful-First RPA~\citep{li2025faithact} uses faithfulness signals
to plan tool actions during inference. Closer to our setting,
MAPO~\citep{yang2026walktalkbridgingreasoningaction} identifies a reasoning–action discrepancy
in tool-using multimodal agents under outcome-based rewards, and
Tool-Use Tax~\citep{zhang2026toolsneedunveilingtooluse} decomposes
prompt formatting, protocol overhead, and execution gain in
LLM agents, finding that protocol cost can offset tool gain under
semantic noise. These works propose corrective training or
inference-time mechanisms; we instead provide a sample-level
empirical characterization of when tool calls fail to expand the
set of problems already solvable without tools.

\subsection{Benchmarks for Agentic Multimodal Reasoning}

Recent benchmarks broaden multimodal evaluation toward agentic
settings that explicitly stress tool use, including agentic
thinking-with-images across diverse tasks~\citep{li2025tir},
long-horizon hybrid tool use across many
sub-domains~\citep{su2026agentvistaevaluatingmultimodalagents}, and knowledge-intensive
geolocation reasoning with expert-annotated stepwise
traces~\citep{geng2026geobrowse}. These benchmarks make tool use
more central to evaluation, but their primary goal is to measure
performance under harder or more process-aware tasks. We ask a
complementary question: is the observed tool-enabled performance
actually attributable to tool access?

\section{Study Design}

\begin{table*}[!t]
  \centering
  \caption{Main math reasoning benchmark results across six representative benchmarks. The bold numbers indicate the best performance on each benchmark, and the underlined numbers indicate the second best.}
  \label{tab:main-math-reason}
  \small
  \setlength{\tabcolsep}{4pt}
  \resizebox{\textwidth}{!}{%
\begin{tabular}{lc | cccccc}
\toprule
Model & Tool & \multicolumn{6}{c}{Math Reasoning} \\
\specialrule{0em}{0pt}{1pt}
\cline{3-8}
\specialrule{0em}{0pt}{1pt}
& & MathVista & MathVerse & MathVision & WeMath & DynaMath & LogicVista \\
\midrule
Qwen2.5-VL   & \ding{55} & 68.3 & 45.6 & 25.6 & 34.6 & 53.3 & 45.9 \\
MM-Eureka    & \ding{55} & \underline{72.6} & -    & 28.1 & 21.8 & -    & 46.3 \\
VL-Rethinker & \ding{55} & \underline{73.7} & -    & \underline{28.4} & 36.3 & -    & 42.7 \\
DeepEyes     & Crop      & 70.1 & \underline{47.3} & 26.6 & \underline{38.9} & 55.0 & 47.7 \\
Thyme        & Code      & 71.9 & 47.3 & 25.5 & 37.5 & 54.1 & 48.3 \\
Thyme        & Tool-Free & 71.7 & 41.6 & 24.9 & 37.5 & 54.3 & 47.4 \\
DeepEyesV2   & General   & 72.5 & \textbf{52.2} & \underline{28.4} & 37.8 & \textbf{57.6} & 48.1 \\
DeepEyesV2   & Tool-Free & 73.8 & 48.6 & 28.5 & \underline{43.1} & \underline{57.9} & \underline{49.2} \\
Pure-Text Reasoner & \ding{55} & \textbf{74.6} & \underline{51.6} & \textbf{29.3} & \textbf{46.3} & 56.8 & \textbf{49.2} \\
\bottomrule
\end{tabular}%
  }
\end{table*}

\subsection{Study Settings}

This section describes the five settings used throughout the paper, the two multimodal agents they are applied to, the benchmarks covered, and how the Pure-Text Reasoner is trained. Three settings test whether tool access expands the set of problems an agent can solve; the remaining two isolate whether behavior depends more on the model-generated tool-call message or on the returned execution result.

\subsection{Comparison Settings}
\label{sec:comparison-setting}

\paragraph{Tool-Enabled Agent.} The trained multimodal agent is run with its default tool configuration, exactly as released. This is the reference point against which all other settings are compared.

\paragraph{Tool-Free Agent.}
The same agent model is run with a prompt that forbids code generation and
instructs text-only reasoning. This is an out-of-distribution diagnostic setting rather than a matched training control: if the released agent's successes depend strongly on executed tools, suppressing tool calls should produce a visible drop.

\begin{figure*}[!t]
  \centering
  \begin{subfigure}{0.49\textwidth}
    \centering
    \includegraphics[width=\textwidth]{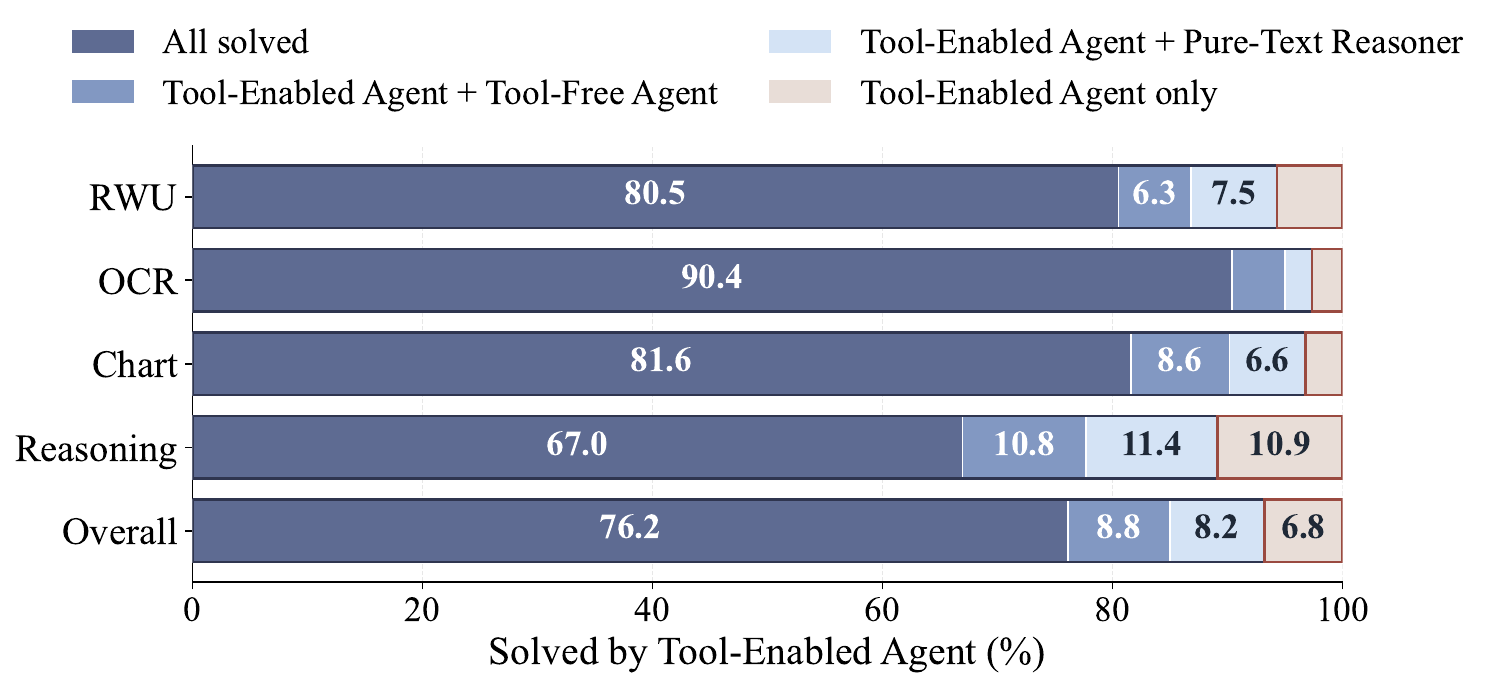}
    \caption{Solved-set coverage for DeepEyesV2.}
    \label{fig:agent-success-coverage-deepeyesv2}
  \end{subfigure}
  \hfill
  \begin{subfigure}{0.49\textwidth}
    \centering
    \includegraphics[width=\textwidth]{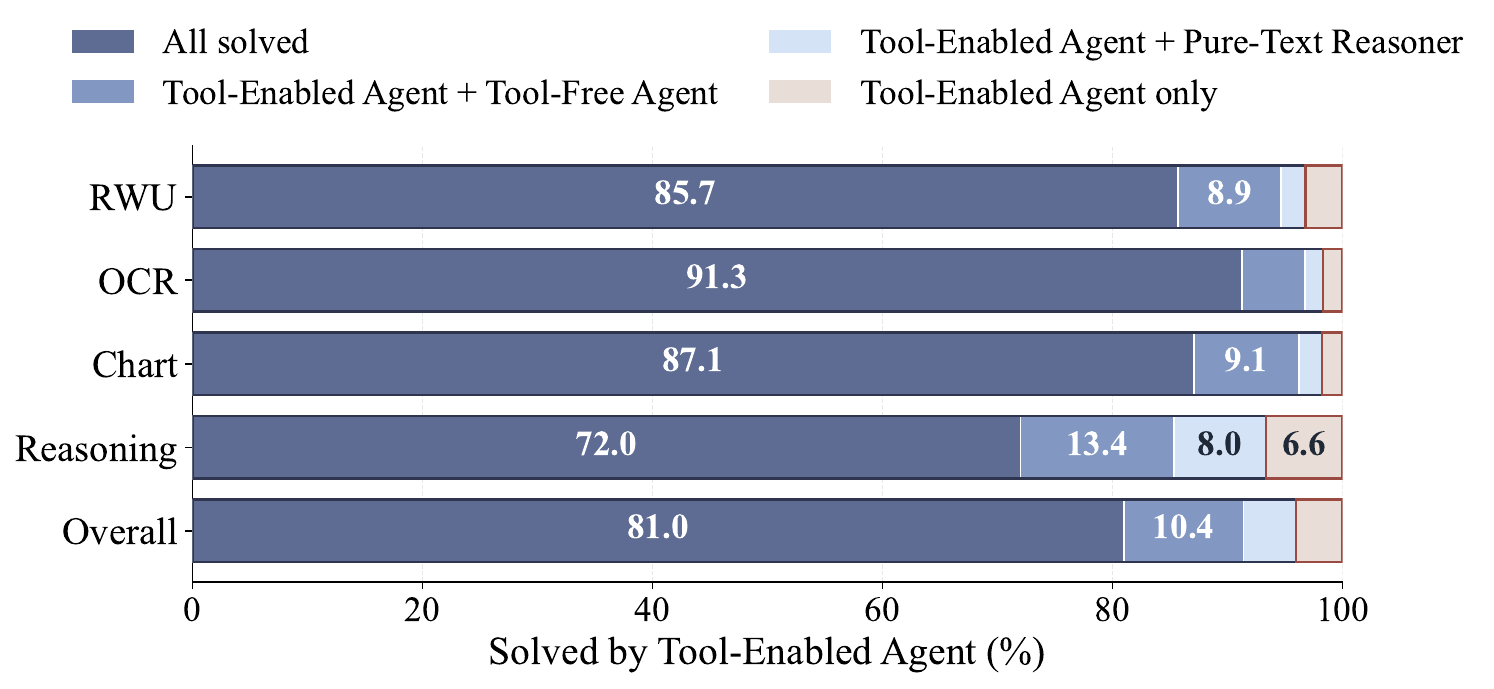}
    \caption{Solved-set coverage for Thyme.}
    \label{fig:agent-success-coverage-thyme}
  \end{subfigure}
  \caption{Solved-set coverage of Tool-Enabled Agents relative to
non-tool settings across task families. Most problems solved by the
Tool-Enabled Agent are also solved by at least one non-tool setting,
while the tool-only solved region remains comparatively small across
all task families. Each bar is normalized over the samples correctly
solved by the corresponding Tool-Enabled Agent. “Tool-Enabled only”
denotes samples missed by both the Tool-Free Agent and the Pure-Text
Reasoner; “All solved” denotes samples solved by all three settings.}
  \label{fig:agent-success-coverage}
\end{figure*}

\paragraph{Pure-Text Reasoner.} A separate baseline trained under a tool-free regime, initialized from the same backbone family and using data drawn from the same source pool as the agents under study. Unlike the Tool-Free Agent, which was trained with tools and has tool access suppressed only at inference time, this model has never seen a tool-calling trajectory. It therefore serves as a tool-free training reference: it estimates what can be achieved from processed data drawn from the same source pool without learning a tool-calling interface. It is not a strict training-procedure control or an upper-bound baseline, but an auxiliary reference for solved-set attribution.

The remaining two settings isolate which part of the tool-use loop a behavior depends on by intervening on already-generated trajectories rather than retraining the model. Both Thyme and DeepEyesV2 invoke tools by generating executable code and folding the returned execution result — an edited image, a numerical value, or parsed text — back into the reasoning chain. In \textit{Tool Format Only}, we post-hoc edit each tool-enabled trajectory by retaining the model-generated tool-call message, including any executable code, and replacing the returned execution result with an empty placeholder before the trajectory re-enters the model for the final answer. In \textit{Tool Result Only}, we do the reverse: the tool-call message is removed from the trajectory, while the corresponding execution result is retained at the position where it would have appeared. Both interventions operate on already-generated trajectories; no retraining is involved. Their contrast reveals whether behavior tracks the call it produces, the result it receives, or both.

We study Thyme and DeepEyesV2. Both generate executable code during reasoning, run it with an external interpreter, and return the result to the trajectory, but differ in their tool operations and training pipelines. Both provide open-source training and inference code, enabling controlled comparisons, and both are competitive on the benchmarks considered. They are not exhaustive, but cover representative realizations of the current ``thinking with images'' paradigm.

\subsection{Benchmarks}

Evaluation spans four families of multimodal tasks, chosen so that
conclusions are not pinned to any single capability profile:
\textit{real-world understanding}~\citep{wu2024v,wang2025divide,wang2026traceable},
\textit{OCR}~\citep{liu2024ocrbench,Li_2024_CVPR},
\textit{chart understanding}~\citep{wang2024charxiv,masry2022chartqa},
and \textit{mathematical reasoning}~\citep{lu2024mathvista,zhang2024mathverse,wang2024measuring,qiao-etal-2025-math,zou2025dynamath,xiao2024logicvista}.

\begin{table*}[!t]
\centering
\fontsize{4.2pt}{5.0pt}\selectfont

\setlength{\heavyrulewidth}{0.08em}
\setlength{\lightrulewidth}{0.05em}
\setlength{\cmidrulewidth}{0.03em}

\caption{Task-family average of tool-call outcome decomposition on
tool-called samples. Across both agents and both non-tool references,
TRR is the dominant outcome, indicating that many tool-called samples
are already solved correctly without tools, while TER remains
comparatively small. This suggests that tool calls more often preserve
existing correctness than convert previously incorrect predictions into
correct ones. Rates are computed against both the Tool-Free Agent and
the Pure-Text Reasoner.}
\label{tab:tool-call-benefit-task-avg}

\setlength{\tabcolsep}{2.8pt}
\renewcommand{\arraystretch}{1.12}
\setlength{\aboverulesep}{0pt}
\setlength{\belowrulesep}{0pt}

\resizebox{\textwidth}{!}{%
\begin{tabular}{c>{\centering\arraybackslash}m{2.7cm}cccccccc}
\toprule

\multirow{2}{*}{\textbf{Model}}
& \multirow{2}{*}{\textbf{Task}}
& \multicolumn{4}{c}{\textbf{vs Tool-Free Agent}}
& \multicolumn{4}{c}{\textbf{vs Pure-Text Reasoner}} \\

\cmidrule(lr){3-6}
\cmidrule(lr){7-10}

& & \textbf{TRR} & \textbf{TFR} & \textbf{TER} & \textbf{TIR}
& \textbf{TRR} & \textbf{TFR} & \textbf{TER} & \textbf{TIR} \\

\midrule

\multirow{5}{*}{\textbf{DeepEyesV2}}

& Real-World Understanding
& \cellcolor[HTML]{6E89A4}\textcolor{white}{59.1}
& \cellcolor[HTML]{B6C6D5}\textcolor{black}{24.8}
& \cellcolor[HTML]{DCE3EB}\textcolor{black}{9.0}
& \cellcolor[HTML]{DCE3EB}\textcolor{black}{7.2}
& \cellcolor[HTML]{6E89A4}\textcolor{white}{59.9}
& \cellcolor[HTML]{B6C6D5}\textcolor{black}{22.8}
& \cellcolor[HTML]{DCE3EB}\textcolor{black}{8.2}
& \cellcolor[HTML]{DCE3EB}\textcolor{black}{9.2} \\

\cmidrule(lr){2-10}

& OCR
& \cellcolor[HTML]{587896}\textcolor{white}{69.7}
& \cellcolor[HTML]{B6C6D5}\textcolor{black}{20.4}
& \cellcolor[HTML]{DCE3EB}\textcolor{black}{3.7}
& \cellcolor[HTML]{DCE3EB}\textcolor{black}{6.2}
& \cellcolor[HTML]{587896}\textcolor{white}{68.0}
& \cellcolor[HTML]{B6C6D5}\textcolor{black}{19.1}
& \cellcolor[HTML]{DCE3EB}\textcolor{black}{5.3}
& \cellcolor[HTML]{DCE3EB}\textcolor{black}{7.5} \\

\cmidrule(lr){2-10}

& Chart
& \cellcolor[HTML]{587896}\textcolor{white}{68.1}
& \cellcolor[HTML]{B6C6D5}\textcolor{black}{17.6}
& \cellcolor[HTML]{DCE3EB}\textcolor{black}{7.4}
& \cellcolor[HTML]{DCE3EB}\textcolor{black}{6.8}
& \cellcolor[HTML]{587896}\textcolor{white}{66.6}
& \cellcolor[HTML]{B6C6D5}\textcolor{black}{16.3}
& \cellcolor[HTML]{DCE3EB}\textcolor{black}{8.9}
& \cellcolor[HTML]{DCE3EB}\textcolor{black}{8.1} \\

\cmidrule(lr){2-10}

& Reasoning
& \cellcolor[HTML]{99AEC3}\textcolor{black}{39.8}
& \cellcolor[HTML]{99AEC3}\textcolor{black}{37.1}
& \cellcolor[HTML]{C9D5E0}\textcolor{black}{11.4}
& \cellcolor[HTML]{C9D5E0}\textcolor{black}{11.6}
& \cellcolor[HTML]{99AEC3}\textcolor{black}{40.1}
& \cellcolor[HTML]{99AEC3}\textcolor{black}{36.5}
& \cellcolor[HTML]{C9D5E0}\textcolor{black}{11.1}
& \cellcolor[HTML]{C9D5E0}\textcolor{black}{12.3} \\

\cmidrule(lr){2-10}

& \textbf{Overall}
& \cellcolor[HTML]{83A0B7}\textcolor{black}{52.3}
& \cellcolor[HTML]{B6C6D5}\textcolor{black}{29.0}
& \cellcolor[HTML]{DCE3EB}\textcolor{black}{9.3}
& \cellcolor[HTML]{DCE3EB}\textcolor{black}{9.4}
& \cellcolor[HTML]{83A0B7}\textcolor{black}{52.0}
& \cellcolor[HTML]{B6C6D5}\textcolor{black}{28.0}
& \cellcolor[HTML]{DCE3EB}\textcolor{black}{9.6}
& \cellcolor[HTML]{C9D5E0}\textcolor{black}{10.4} \\

\midrule

\multirow{5}{*}{\textbf{Thyme}}

& Real-World Understanding
& \cellcolor[HTML]{587896}\textcolor{white}{71.0}
& \cellcolor[HTML]{B6C6D5}\textcolor{black}{20.0}
& \cellcolor[HTML]{DCE3EB}\textcolor{black}{2.7}
& \cellcolor[HTML]{DCE3EB}\textcolor{black}{6.3}
& \cellcolor[HTML]{587896}\textcolor{white}{66.7}
& \cellcolor[HTML]{B6C6D5}\textcolor{black}{18.8}
& \cellcolor[HTML]{DCE3EB}\textcolor{black}{7.1}
& \cellcolor[HTML]{DCE3EB}\textcolor{black}{7.5} \\

\cmidrule(lr){2-10}

& OCR
& \cellcolor[HTML]{426986}\textcolor{white}{83.7}
& \cellcolor[HTML]{C9D5E0}\textcolor{black}{13.2}
& \cellcolor[HTML]{EFF2F6}\textcolor{black}{0.5}
& \cellcolor[HTML]{DCE3EB}\textcolor{black}{2.7}
& \cellcolor[HTML]{426986}\textcolor{white}{81.0}
& \cellcolor[HTML]{C9D5E0}\textcolor{black}{10.0}
& \cellcolor[HTML]{DCE3EB}\textcolor{black}{3.2}
& \cellcolor[HTML]{DCE3EB}\textcolor{black}{5.9} \\

\cmidrule(lr){2-10}

& Chart
& \cellcolor[HTML]{6E89A4}\textcolor{white}{62.5}
& \cellcolor[HTML]{99AEC3}\textcolor{black}{31.2}
& \cellcolor[HTML]{EFF2F6}\textcolor{black}{0.0}
& \cellcolor[HTML]{DCE3EB}\textcolor{black}{6.2}
& \cellcolor[HTML]{99AEC3}\textcolor{black}{37.5}
& \cellcolor[HTML]{C9D5E0}\textcolor{black}{12.5}
& \cellcolor[HTML]{B6C6D5}\textcolor{black}{25.0}
& \cellcolor[HTML]{B6C6D5}\textcolor{black}{25.0} \\

\cmidrule(lr){2-10}

& Reasoning
& \cellcolor[HTML]{B6C6D5}\textcolor{black}{31.7}
& \cellcolor[HTML]{83A0B7}\textcolor{black}{43.1}
& \cellcolor[HTML]{B6C6D5}\textcolor{black}{22.0}
& \cellcolor[HTML]{DCE3EB}\textcolor{black}{3.3}
& \cellcolor[HTML]{B6C6D5}\textcolor{black}{30.9}
& \cellcolor[HTML]{99AEC3}\textcolor{black}{37.4}
& \cellcolor[HTML]{B6C6D5}\textcolor{black}{22.8}
& \cellcolor[HTML]{DCE3EB}\textcolor{black}{8.9} \\

\cmidrule(lr){2-10}

& \textbf{Overall}
& \cellcolor[HTML]{587896}\textcolor{white}{71.3}
& \cellcolor[HTML]{B6C6D5}\textcolor{black}{20.3}
& \cellcolor[HTML]{DCE3EB}\textcolor{black}{4.5}
& \cellcolor[HTML]{DCE3EB}\textcolor{black}{4.0}
& \cellcolor[HTML]{587896}\textcolor{white}{67.9}
& \cellcolor[HTML]{B6C6D5}\textcolor{black}{17.0}
& \cellcolor[HTML]{DCE3EB}\textcolor{black}{7.8}
& \cellcolor[HTML]{DCE3EB}\textcolor{black}{7.2} \\

\bottomrule
\end{tabular}%
}

\end{table*}

\subsection{Training the Pure-Text Reasoner}

The Pure-Text Reasoner is a diagnostic tool-free reference rather than
an upper-bound model. It estimates which successes are attainable from
the same source pool without learning a tool-use protocol. To align it
with DeepEyesV2, we initialize from Qwen2.5-VL-7B and train on processed
data derived from the DeepEyesV2 source pool.

Training has two stages. In SFT, we extract 53K examples from the DeepEyesV2 SFT data. Because the original rationales include tool-call steps, which are incompatible with a pure-text setting, we use Qwen2.5-VL-72B to convert them into text-based reasoning traces without tool use. In RL, we apply a pass@$k$ filter to the DeepEyesV2 RL data for computational feasibility, retaining 30K examples with $1 \le n \le k-1$ correct sampled responses, and train with GRPO~\citep{shao2024deepseekmathpushinglimitsmathematical} using final-answer correctness as the reward. We use Qwen3-235B-A22B-Instruct-2507~\citep{yang2025qwen3technicalreport} to judge final-answer correctness for the reward. Additional details are
provided in Appendix~\ref{Training Details}.

\section{Experiment Results}

\subsection{Tool Access Does Not Consistently Expand Solved Sets}
\label{sec:no-gains}

If tool access provides a robust capability gain, we would expect it to improve accuracy on tool-relevant benchmarks or reduce the generated-token cost needed to reach comparable answers. In our experiments, neither pattern appears consistently.

\begin{figure*}[!t]
  \centering
  \makebox[\textwidth][l]{%
  \hspace*{-0.04\textwidth}%
  \begin{subfigure}[t]{0.36\textwidth}
    \centering
    \includegraphics[height=4cm]{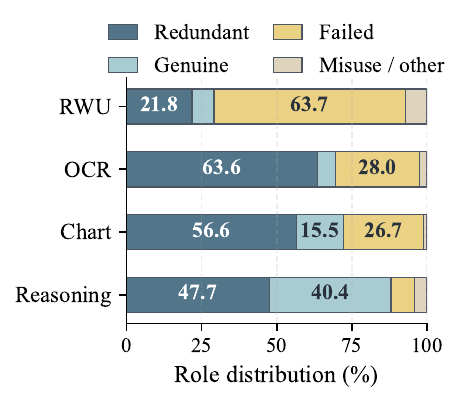}
    \caption{DeepEyesV2.}
    \label{fig:tool-process-deepeyesv2}
  \end{subfigure}
  \hspace{0\textwidth}
  \begin{subfigure}[t]{0.36\textwidth}
    \centering
    \includegraphics[height=4cm]{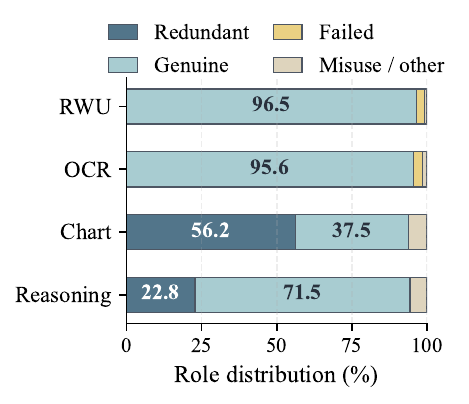}
    \caption{Thyme.}
    \label{fig:tool-process-thyme}
  \end{subfigure}
  \hspace{0.02\textwidth}
  \begin{subfigure}[t]{0.22\textwidth}
    \centering
    \includegraphics[height=4cm]{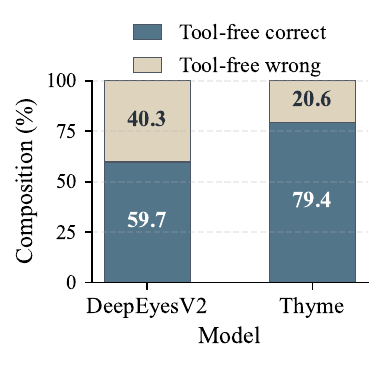}
    \caption{\textit{Genuine contribution} subset.}
    \label{fig:tool-process-source}
  \end{subfigure}%
  }

  \caption{
Process-level analysis of tool use. Tool calls often reflect
redundant confirmation, failed contribution, or contributions to
samples that are already solvable without tools, rather than clear
capability expansion. The first two subfigures show tool-use role
distributions across task families. The third subfigure further shows
that even among trajectories labeled as \textit{Genuine contribution},
most samples are already solved by the Tool-Free Agent. Role-distribution sample sizes are:
DeepEyesV2 --- 2195/3275/7499/15175 and Thyme --- 255/410/16/123
for RWU/OCR/Chart/Reasoning, respectively. Small Thyme subsets,
especially Chart and Reasoning, should be interpreted descriptively.
}
  \label{fig:tool-process-analysis}
\end{figure*}

\paragraph{Aggregate accuracy.}
Across perception and reasoning benchmarks
(Tables~\ref{tab:main-perception} and~\ref{tab:main-math-reason}),
the Tool-Enabled Agents are competitive and occasionally top an
individual benchmark, but no consistent advantage emerges across task
families. Moreover, disabling tools does not lead to a systematic collapse in performance:
the Tool-Free variants often remain close to, and occasionally surpass, their
tool-enabled counterparts, suggesting that many successes do not depend on
executed tool use. On several benchmarks --- spanning OCR, charts, and math
reasoning --- the Pure-Text Reasoner matches or surpasses them, despite being trained on processed data derived from the same source pool
without any tool-calling mechanism. Thus, aggregate accuracy does not show a consistent advantage
for tool-enabled inference.

\paragraph{Solved-set composition.}
Aggregate scores can mask which problems an agent actually solves. Two
settings can land at the same accuracy by solving largely the same
subset, or by solving disjoint ones --- and the difference matters: only
the latter would suggest tools open up problems otherwise out of reach.
This analysis is intended for attribution rather than causal estimation or model ranking: it asks whether tool-enabled successes are also attainable without tool execution under comparable non-tool references. To tell these apart, we
condition on the samples each Tool-Enabled Agent solves, and check how
many of them are also solved by the Tool-Free Agent or the Pure-Text
Reasoner.


\begin{table*}[!t]
\centering
\small
\setlength{\tabcolsep}{4pt}
\renewcommand{\arraystretch}{1.08}
\caption{Per-sample correctness agreement (\%) with the Tool-Enabled Agent, computed as the proportion of samples on which the two settings yield the same correctness outcome (both correct or both incorrect). Higher values indicate that the comparison setting more closely matches the correctness pattern of the Tool-Enabled Agent. Parenthesized values report the difference from the Tool-Free Agent.}
\label{tab:module3-category-agreement}
\resizebox{\textwidth}{!}{%
\begin{tabular}{lcccccc}
\toprule
\multirow{2}{*}{Task}
& \multicolumn{3}{c}{DeepEyesV2}
& \multicolumn{3}{c}{Thyme} \\
\cmidrule(lr){2-4} \cmidrule(lr){5-7}
& Tool-Free
& Result Only
& Format Only
& Tool-Free
& Result Only
& Format Only \\
\midrule
Real-World Understanding & 82.9 & 82.8 {\scriptsize\textcolor{red!70!black}{(-0.1)}} & \textbf{90.0} {\scriptsize\textcolor{blue!70!black}{(+7.1)}} & 92.4 & \textbf{99.4} {\scriptsize\textcolor{blue!70!black}{(+7.0)}} & 99.2 {\scriptsize\textcolor{blue!70!black}{(+6.8)}} \\
OCR                      & 90.7 & 92.8 {\scriptsize\textcolor{blue!70!black}{(+2.1)}} & \textbf{93.9} {\scriptsize\textcolor{blue!70!black}{(+3.2)}} & 95.9 & \textbf{99.0} {\scriptsize\textcolor{blue!70!black}{(+3.2)}} & 96.5 {\scriptsize\textcolor{blue!70!black}{(+0.6)}} \\
Chart                    & 83.3 & 85.8 {\scriptsize\textcolor{blue!70!black}{(+2.5)}} & \textbf{90.8} {\scriptsize\textcolor{blue!70!black}{(+7.5)}} & 93.0 & \textbf{96.3} {\scriptsize\textcolor{blue!70!black}{(+3.3)}} & 95.9 {\scriptsize\textcolor{blue!70!black}{(+2.9)}} \\
Reasoning                & 76.8 & 77.7 {\scriptsize\textcolor{blue!70!black}{(+0.8)}} & \textbf{82.9} {\scriptsize\textcolor{blue!70!black}{(+6.0)}} & 84.8 & \textbf{89.6} {\scriptsize\textcolor{blue!70!black}{(+4.9)}} & 88.8 {\scriptsize\textcolor{blue!70!black}{(+4.0)}} \\
\bottomrule
\end{tabular}%
}
\end{table*}

Figure~\ref{fig:agent-success-coverage} shows the breakdown by task family.
Across all benchmarks, 93.2\% of the problems DeepEyesV2 solves with tools
are also solved by at least one non-tool setting; for Thyme the figure is
96.0\%. This overlap is not driven solely by the separately trained Pure-Text Reasoner: Appendix Table~\ref{tab:app-solved-set-benchmark} reports the benchmark-level overlap with the released Tool-Free Agent separately and shows the same qualitative pattern. The pattern holds in every task family. The tool-only solved set ---
samples correctly answered by the Tool-Enabled Agent but missed by both
non-tool settings --- remains small overall, with most task-family values
in the single digits or low double digits.

\paragraph{No compensating efficiency.}
Tool-enabled inference also does not
consistently reduce generated-token cost. Across task families,
Tool-Enabled DeepEyesV2 generates 445 tokens per sample on average,
compared with 413 for the Tool-Free Agent and 275 for the Pure-Text
Reasoner; for Thyme, the corresponding numbers are 272, 259, and
275. Detailed token counts are reported in Appendix Table~\ref{tab:app-token-cost-benchmark}.

\paragraph{Implications.}
These results show why aggregate benchmark scores are insufficient for
evaluating tool use. Small score differences need not imply that tools
open new solvable regions; they may instead reflect changes near an
existing decision boundary. This motivates a more local question: when an
agent invokes a tool, what does the call actually contribute?


\subsection{Tool Calls Confirm More Than They Repair}

\label{sec:confirm-not-repair}
Section~\ref{sec:no-gains} showed that tool access does not consistently
expand the solved set. We next ask what role tool calls play when they occur, using two complementary views: outcome counts, which show whether tool calling preserves, repairs, or harms correctness, and trajectory attribution, which asks whether the tool output supplies missing information or merely confirms the model's prior inference. Across both views, tool calls are dominated by redundant confirmation and failed repair rather than genuine error correction.

\paragraph{Tool-call outcome decomposition.}
On tool-called samples, we compare the Tool-Enabled Agent with a
non-tool reference and partition outcomes into four mutually exclusive
cases: \textit{TRR}, where both settings are correct; \textit{TFR},
where both are wrong; \textit{TER}, where the non-tool reference is
wrong but tool calling makes the answer correct; and \textit{TIR}, where
the non-tool reference is correct but tool calling makes the answer
wrong. Rates are normalized over tool-called samples and sum to
100\%. To avoid pinning conclusions to a single reference, we compute
them against both the Tool-Free Agent and the Pure-Text Reasoner.


\begin{table*}[t]
\centering
\small
\setlength{\tabcolsep}{3.5pt}
\renewcommand{\arraystretch}{1.08}
\caption{Performance comparison across tool-usage modes on Real-World Understanding, OCR, and Chart benchmarks.}
\label{tab:mode-ablation}
\resizebox{\textwidth}{!}{%
\begin{tabular}{llccccccccc}
\toprule
\multirow{2}{*}{Model} & \multirow{2}{*}{Mode}
& \multicolumn{4}{c}{Real-World Understanding}
& \multicolumn{2}{c}{OCR}
& \multicolumn{3}{c}{Chart} \\
\cmidrule(lr){3-6} \cmidrule(lr){7-8} \cmidrule(lr){9-11}
&
& V* Bench & HRBench 4K & HRBench 8K & Tree Bench
& OCR Bench & SEED 2 PLUS
& CharXiv descriptive & CharXiv reasoning & Chart QA \\
\midrule
\multirow{3}{*}{DeepEyesV2}
& Tool-Enabled Agent & 82.20 & 76.13 & 72.25 & 41.48 & 814 & 69.78 & 75.68 & 45.90 & 88.08 \\
& Tool Format Only & 83.77 & 76.00 & 70.88 & 39.01 & 805 & 69.78 & 76.05 & 44.70 & 87.08 \\
& Tool Result Only & 81.68 & 73.50 & 68.75 & 34.07 & 828 & 69.87 & 73.78 & 45.90 & 80.24 \\
\midrule
\multirow{3}{*}{Thyme}
& Tool-Enabled Agent & 82.72 & 77.00 & 72.50 & 38.77 & 865 & 70.22 & 73.30 & 43.80 & 87.68 \\
& Tool Format Only & 83.25 & 77.25 & 72.13 & 39.26 & 865 & 70.05 & 73.15 & 42.70 & 87.08 \\
& Tool Result Only & 83.25 & 77.50 & 72.38 & 39.26 & 879 & 70.36 & 73.00 & 42.20 & 87.56 \\
\midrule
Pure-Text Reasoner & -- & 83.25 & 77.25 & 72.50 & 42.22 & 888 & 70.66 & 77.45 & 46.20 & 83.44 \\
\bottomrule
\end{tabular}%
}
\end{table*}


The dominant case, by a wide margin, is \textit{TRR}
(Table~\ref{tab:tool-call-benefit-task-avg}). Against the Tool-Free
Agent, overall \textit{TRR} reaches 52.3\% for DeepEyesV2 and 71.3\%
for Thyme, while \textit{TER} stays below 10\% for both. Using the
Pure-Text Reasoner as the baseline gives the same picture. Most tool
calls, in other words, occur on samples that a non-tool reference
already answers correctly.

The four rates are not evenly distributed across task families. On
reasoning, the picture flips: \textit{TRR} drops sharply, and
\textit{TFR} climbs to 37.1\% for DeepEyesV2 and 43.1\% for Thyme.
Tools are still being called on these samples, but many calls fail to
convert incorrect predictions into correct ones. This is precisely where
the outcome decomposition is most diagnostic: it shows that tool calls
often do not repair the errors they are expected to address. The full
benchmark-level decomposition is reported in Appendix
Table~\ref{tab:app-tool-call-benefit-full}.

\paragraph{Process attribution.}
Outcome counts cannot separate "tool helped" from "tool happened to coincide with a correct answer". For that, we need trajectory-level attribution. We use Qwen3-VL-30B~\citep{bai2025qwen3vltechnicalreport} as the judge model, with a manual audit on a random
40\% subset, to label each tool-using trajectory along three axes: whether the tool output was novel relative to the model's pre-tool reasoning, whether it was reliable, and whether the model integrated it into the final answer. The label combinations group naturally into four roles: \textit{Redundant confirmation}, where the output mainly reaffirms what the model already believed; \textit{Genuine contribution}, where the tool supplies new and correct information that the model uses; \textit{Failed / non-contributory}, where the output is unusable; and \textit{Misuse / other}, for incorrect integration and rare combinations. The full prompt and benchmark-level distributions are in Appendix Tables~\ref{tab:appendix-tool-process-deepeyesv2} and~\ref{tab:appendix-tool-process-thyme}.

What the judge sees (Figure~\ref{fig:tool-process-analysis}) lines up with what the outcome decomposition implies. For DeepEyesV2, \textit{Redundant confirmation} is the largest role on OCR (63.6\%) and chart (56.6\%) tasks; on real-world understanding, \textit{Failed / non-contributory} takes over (63.7\%); even on reasoning, \textit{Redundant confirmation} still leads at 47.7\%, with \textit{Genuine contribution} rising to 40.4\%. Thyme appears to assign more mass to \textit{Genuine contribution}, although this pattern should be interpreted with care because several Thyme task-family estimates are based on relatively few tool-called samples. As we show next, even this label often corresponds to samples that are already solved without tools.

\paragraph{Many “genuine contribution” cases are already solvable without tools.}

This leaves one place where tool calls might still be doing real work: the trajectories the judge labels as \textit{Genuine contribution}. Figure~\ref{fig:tool-process-source} examines them. Of those samples, 59.7\% (DeepEyesV2) and 79.4\% (Thyme) were already solved by the Tool-Free Agent. The trajectories that look most substantive --- novel information, correctly integrated --- often occur on problems the model could already
solve, especially for Thyme. The tool calls that the judge identifies as
informative are therefore frequently not necessary for final correctness.

\paragraph{Implications.}
The outcome counts and trajectory annotations converge on the same pattern: tool calls frequently occur on samples that are already solvable, while samples that require correction often remain incorrect. This does not match the behavior of a model that has learned to reach for a tool when its own knowledge runs out. In this sense, much of the observed tool use looks like a learned reflex — a pattern of calling that is weakly connected to whether a tool is actually needed.

\subsection{Call Format or Returned Result: What Drives Tool-Conditioned Behavior?}
\label{sec:format-or-content}

\begin{table*}[t]
\centering
\small
\setlength{\tabcolsep}{4pt}
\renewcommand{\arraystretch}{1.08}
\caption{Performance comparison across tool-usage modes on reasoning benchmarks.}
\label{tab:mode-ablation-reasoning}
\resizebox{\textwidth}{!}{%
\begin{tabular}{llcccccc}
\toprule
Model & Mode & MathVista & MathVerse & MathVision & WeMath & DynaMath & LogicVista \\
\midrule
\multirow{3}{*}{DeepEyesV2}
& Tool-Enabled Agent & 72.50 & 52.21 & 28.39 & 37.81 & 57.60 & 48.10 \\
& Tool Format Only & 71.20 & 47.99 & 27.27 & 36.19 & 57.53 & 47.18 \\
& Tool Result Only & 69.70 & 40.20 & 28.06 & 29.90 & 48.42 & 43.37 \\
\midrule
\multirow{3}{*}{Thyme}
& Tool-Enabled Agent & 71.90 & 47.26 & 25.49 & 37.52 & 54.11 & 48.32 \\
& Tool Format Only & 71.30 & 46.70 & 26.18 & 38.67 & 53.45 & 50.56 \\
& Tool Result Only & 70.90 & 47.39 & 25.26 & 37.43 & 54.21 & 47.20 \\
\midrule
Pure-Text Reasoner & -- & 74.60 & 51.62 & 29.31 & 46.29 & 56.81 & 49.22 \\
\bottomrule
\end{tabular}%
}
\end{table*}

We next ask which part of the tool-use loop drives tool-conditioned
behavior: the model-generated tool-call message, the returned execution
result, or both. As defined in Section~\ref{sec:comparison-setting}, we
intervene on already-generated trajectories: \textit{Tool Format Only}
keeps the call message but suppresses the result, while \textit{Tool Result
Only} keeps the result but removes the call message. We then test which
intervention better preserves each agent's correctness pattern.

\paragraph{Behavioral agreement.}
For two settings $A$ and $B$, we define their per-sample correctness agreement as $\mathrm{Agree}(A, B) = \frac{1}{N}\sum_{i=1}^{N} \mathbf{1}[c^A_i = c^B_i]$, where $c^A_i, c^B_i \in \{0, 1\}$ are correctness indicators on sample $i$. Higher values indicate that the two settings make the same correctness judgments on more samples. Table~\ref{tab:module3-category-agreement} reports per-sample correctness agreement (\%) between each intervention and the Tool-Enabled Agent — that is, the proportion of samples on which the two settings yield the same correctness outcome — alongside the Tool-Free Agent as a reference floor. The two agents land on opposite sides. 

For DeepEyesV2, \textit{Tool Format Only} is closer to the Tool-Enabled Agent than the Tool-Free Agent on every task family, and is the closest of the three comparison settings overall, with agreement gains of $+7.1$, $+7.5$, and $+6.0$ points on real-world understanding, chart, and reasoning. In other words, keeping the tool-call format while removing the returned result disturbs DeepEyesV2 less than the reverse intervention. The generated call format accounts for much of the behavioral pattern attributed to tool use.

Thyme shows the opposite pattern. \textit{Tool Result Only} achieves the highest agreement on all four task families, with gains of $+3.2$ to $+7.0$ points over the Tool-Free Agent. \textit{Tool Format Only} also tracks the Tool-Enabled Agent closely, but remains slightly below \textit{Tool Result Only}. Thus, Thyme’s behavior is better preserved by the returned execution result than by the calling pattern alone.

\paragraph{Final-task performance.}

The behavioral picture is consistent with how each intervention affects task scores (Tables~\ref{tab:mode-ablation} and~\ref{tab:mode-ablation-reasoning}). For DeepEyesV2, performance under \textit{Tool Format Only} remains close to the Tool-Enabled Agent and even slightly improves on V* Bench and CharXiv descriptive; \textit{Tool Result Only}, by contrast, produces visible drops on HRBench, TreeBench, and ChartQA. For Thyme, both interventions stay close to the Tool-Enabled Agent, and \textit{Tool Result Only} matches or surpasses it on OCRBench and HRBench 4K/8K. In both agents, however, the full tool-use pipeline does not deliver a consistent benchmark advantage over the corresponding ablations.

\paragraph{Implications.}
The two agents preserve tool-conditioned behavior through different
parts of the tool loop: DeepEyesV2 through call format, and Thyme
through returned results. This divergence is informative, but the
overall picture remains similar: the full pipeline shows no consistent advantage over either ablation and does not resolve the limited solved-set expansion in Section~\ref{sec:no-gains}. Tool training therefore appears to shape agent conditioning without reliably expanding what they can
solve.

\FloatBarrier

\section{Conclusion}
It has become natural to assume that adding tools to a multimodal
agent, and training it to call them, makes the agent more capable. Our
results suggest a more careful distinction. Across two trained agents, four task families, and analyses of correctness, process, generated-token cost, and mechanism, we find little evidence that tool access consistently expands the solved set relative to comparable non-tool settings. Tool calls often occur on samples already solved without tools, fail to repair incorrect predictions, and are preserved by different parts of the tool-use loop in different agents. These patterns point to a gap between learning to call tools and learning when tool outputs change what can be solved: final-answer correctness alone does not distinguish tools that supply missing information from those that merely confirm an available answer. Evaluation and training signals should therefore credit tools for the information they add, not merely for their presence in successful trajectories; otherwise, scaling tool access may reinforce tool-calling frequency rather than meaningful tool use.

\section*{Limitations}
A few boundaries on this study are worth flagging.
First, our analysis centers on two representative ``thinking with
images'' agents and four task families. The agent ecosystem continues
to grow, so the patterns we report should be read as characterizing
this slice of the field rather than the field as a whole. The
Pure-Text Reasoner also has a limited role: although it uses the same
backbone scale and source data pool and contains no tool-calling
trajectories, it is separately trained with re-annotation and filtering.
We therefore use it as a diagnostic reference for solved-set
attribution, not as evidence that tool-free training is generally
superior.

Second, our framework characterizes tool use at three levels:
aggregate performance and solved-set composition, per-sample outcomes
with process attribution, and mechanism-level ablations. Complementary
perspectives, such as user-perceived helpfulness, multi-turn
interaction, or domain-specific deployment, could further enrich this
view.

Third, the process-level attribution relies on a strong language model as a judge, although a 40\% manual audit supports the major role distinctions. Alternative attribution paradigms, such as multi-judge ensembles or fine-grained human studies, could provide additional triangulation. Nevertheless, the qualitative patterns are consistent across all four analytical dimensions, suggesting that they are not artifacts of any single judge.

\nocite{*}
\bibliography{custom}

\clearpage
\appendix

\section{Training Data for the Pure-Text Reasoner}
\label{Training Details}
To keep the Pure-Text Reasoner aligned with the multimodal agents
under comparison, we construct its training data from the DeepEyesV2
training source pool. The resulting model is intended as a diagnostic
tool-free reference rather than a stronger competitor or an upper-bound
model.

\paragraph{SFT data.}
The original DeepEyesV2 SFT set contains 65K examples, including both
text-only reasoning data and visual reasoning data. Since our study
focuses on multimodal tool use, we remove the text-only examples and
retain only the visual reasoning data. The original visual rationales
contain many tool-calling trajectories, which are incompatible with the
training objective of the Pure-Text Reasoner. We therefore re-annotate
these examples to remove tool-call segments and retain only text-based
reasoning traces.

DeepEyesV2 uses strong closed-source models such as Gemini 2.5 Pro,
GPT-4o, and Claude Sonnet 4 for SFT data annotation. Due to resource
constraints, we perform the re-annotation with our locally deployed
Qwen2.5-VL-72B model. This choice may introduce differences in
annotation style and quality, but it does not introduce tool-calling
trajectories or execution results into the Pure-Text Reasoner.

\paragraph{RL data.}
The original DeepEyesV2 RL set contains 82K examples. We use the same
RL training configuration as DeepEyesV2 where possible, but full-scale
training on the complete RL set is computationally expensive in our
environment: with the full 82K examples, one training step takes about
16 minutes, and one epoch requires approximately 634 steps, or about
seven days. We therefore apply a pass@$k$ filter and retain 30K
examples for RL training. This reduces one epoch to approximately 235
steps, or about 2.6 days.

We acknowledge that this setup does not exactly reproduce the original
DeepEyesV2 training conditions. However, our goal is not to construct a
stronger training set or to use the Pure-Text Reasoner for model ranking.
Instead, it serves as a diagnostic reference for estimating which
tool-enabled successes are also attainable under a tool-free reasoning
path using data from the same source pool. The main conclusion is also
supported by analyses that do not rely on the Pure-Text Reasoner training
pipeline, including the Tool-Free counterpart and the Tool Format Only
and Tool Result Only ablations. These results similarly show that the
full tool-use loop does not consistently yield additional capability
beyond its component signals.




\section{Benchmarks and Evaluation Metrics}

Following the evaluation settings of DeepEyesV2 and Thyme, we evaluate
the compared models on a broad set of multimodal benchmarks covering
real-world understanding, OCR, chart understanding, and mathematical
reasoning. This benchmark selection is intended to avoid drawing
conclusions from a single task type or capability profile.

\paragraph{Real-world understanding.}
For real-world understanding, we evaluate on V*Bench, HRBench, and
TreeBench, which test fine-grained visual perception, high-resolution
image understanding, and structured visual reasoning in natural scenes.

\paragraph{OCR.}
For OCR-related tasks, we evaluate on OCRBench and SEED-Bench-2-Plus.
These benchmarks measure the ability of multimodal models to recognize
and reason over text under diverse visual conditions.

\paragraph{Chart understanding.}
For chart-related reasoning, we evaluate on CharXiv and ChartQA.
CharXiv includes both descriptive and reasoning-oriented chart
understanding tasks, while ChartQA requires answering questions over
visualized data with both perceptual and logical reasoning.

\begin{table*}[t]
\centering
\small
\setlength{\tabcolsep}{4pt}
\renewcommand{\arraystretch}{1.08}
\caption{Benchmark-level solved-set decomposition conditioned on samples
correctly answered by the Tool-Enabled Agent.
$N_{\mathrm{tool\text{-}solved}}$ denotes the number of such samples and
serves as the denominator for all percentages in each row.
\textit{Tool-Enabled only} denotes samples missed by both non-tool
references, while \textit{Also solved without tools} denotes samples
also solved by at least one non-tool setting. The last three columns
decompose this overlap. Summing 'Tool-Enabled only' and 'Also Pure-Text only' gives the share of Tool-Enabled successes missed by the Tool-Free Agent alone; these values help verify that the Section~\ref{sec:no-gains} pattern is not solely driven by the Pure-Text Reasoner.}
\label{tab:app-solved-set-benchmark}
\resizebox{\textwidth}{!}{%
\begin{tabular}{llrrrrrr}
\toprule
Model & Benchmark & $N_{\mathrm{tool\text{-}solved}}$ & Tool-Enabled only & Also solved without tools & Also Tool-Free only & Also Pure-Text only & All three \\
\midrule
DeepEyesV2 & VStarBench & 156 & 3.8 & 96.2 & 1.9 & 4.5 & 89.7 \\
 & HRBench4K & 606 & 4.3 & 95.7 & 5.1 & 5.9 & 84.7 \\
 & HRBench8K & 564 & 3.9 & 96.1 & 6.9 & 3.9 & 85.3 \\
 & TreeBench & 168 & 18.5 & 81.5 & 12.5 & 28.0 & 41.1 \\
 & OCRBench & 814 & 0.7 & 99.3 & 2.9 & 0.9 & 95.5 \\
 & SEEDBench2\_Plus & 1589 & 3.7 & 96.3 & 5.5 & 3.1 & 87.8 \\
 & CharXiv\_descriptive\_val & 3027 & 2.3 & 97.7 & 6.1 & 5.1 & 86.5 \\
 & CharXiv\_reasoning\_val & 437 & 13.3 & 86.7 & 16.5 & 12.6 & 57.7 \\
 & ChartQA\_TEST & 2202 & 2.4 & 97.6 & 10.4 & 7.5 & 79.6 \\
 & MathVista\_MINI & 725 & 2.6 & 97.4 & 8.8 & 5.8 & 82.8 \\
 & MathVerse\_MINI & 2039 & 13.0 & 87.0 & 10.1 & 13.8 & 63.1 \\
 & MathVision & 804 & 25.7 & 74.3 & 17.0 & 17.0 & 40.2 \\
 & WeMath & 1103 & 4.7 & 95.3 & 10.1 & 10.1 & 75.2 \\
 & DynaMath & 2886 & 9.1 & 90.9 & 9.7 & 9.8 & 71.3 \\
 & LogicVista & 215 & 17.7 & 82.3 & 18.1 & 14.0 & 50.2 \\
\midrule
Thyme & VStarBench & 158 & 1.3 & 98.7 & 5.1 & 1.3 & 92.4 \\
 & HRBench4K & 618 & 1.8 & 98.2 & 7.8 & 1.5 & 89.0 \\
 & HRBench8K & 580 & 3.3 & 96.7 & 8.3 & 2.1 & 86.4 \\
 & TreeBench & 157 & 10.8 & 89.2 & 19.7 & 5.7 & 63.7 \\
 & OCRBench & 865 & 0.2 & 99.8 & 3.8 & 0.5 & 95.5 \\
 & SEEDBench2\_Plus & 1599 & 2.5 & 97.5 & 6.3 & 2.2 & 89.0 \\
 & CharXiv\_descriptive\_val & 2930 & 1.7 & 98.3 & 6.7 & 2.0 & 89.6 \\
 & CharXiv\_reasoning\_val & 438 & 6.4 & 93.6 & 21.2 & 9.1 & 63.2 \\
 & ChartQA\_TEST & 2183 & 1.0 & 99.0 & 9.9 & 0.6 & 88.5 \\
 & MathVista\_MINI & 725 & 2.2 & 97.8 & 9.0 & 4.3 & 84.6 \\
 & MathVerse\_MINI & 1862 & 5.8 & 94.2 & 13.2 & 8.3 & 72.8 \\
 & MathVision & 756 & 19.8 & 80.2 & 19.8 & 13.8 & 46.6 \\
 & WeMath & 1136 & 5.0 & 95.0 & 10.9 & 8.6 & 75.4 \\
 & DynaMath & 2711 & 4.8 & 95.2 & 13.4 & 6.8 & 75.0 \\
 & LogicVista & 216 & 14.4 & 85.6 & 19.0 & 9.7 & 56.9 \\
\bottomrule
\end{tabular}%
}
\end{table*}

\paragraph{Mathematical reasoning.}
For mathematical reasoning, we evaluate on MathVista, MathVerse,
MathVision, WeMath, DynaMath, and LogicVista. These benchmarks cover
visual mathematical reasoning, diagram understanding, dynamic or
multi-step mathematical inference, and logical reasoning in visual
contexts.

All evaluations are conducted using VLMEvalKit~\citep{duan2025vlmevalkitopensourcetoolkitevaluating}, following the evaluation
framework used by DeepEyesV2 and Thyme. For benchmarks that require
model-based answer judgment, we use GPT-4o~\citep{openai2024gpt4ocard} as the judge model across all settings to ensure a consistent evaluation protocol. 

\paragraph{Repeated evaluation stability.}
We repeated the benchmark evaluations under the same evaluation protocol. Across repeated runs, the reported accuracies or normalized scores varied by at most 1.3 points, and the qualitative patterns used in our analysis were stable. Specifically, no consistent aggregate advantage, small tool-only solved regions, and close agreement between full tool use and trajectory ablations hold across runs.

\section{Solved-Set Analysis Details}
\label{app:solved-set-details}

Figure~\ref{fig:agent-success-coverage} in the main text provides a task-family
view of whether tool-enabled inference expands the set of problems an
agent can solve. Here we provide the corresponding benchmark-level
decomposition. Instead of only asking whether a Tool-Enabled Agent
achieves a high aggregate score, we ask a more concrete question for
each benchmark: among the examples it answers correctly, how many are
actually new successes that neither non-tool reference can solve?

For each benchmark, we take the samples correctly answered by the
Tool-Enabled Agent as the conditioned set. We denote the size of this
set as $N_{\mathrm{tool\text{-}solved}}$, which serves as the denominator
for all percentages in Table~\ref{tab:app-solved-set-benchmark}. The column
\textit{Tool-Enabled only} reports the proportion of these correctly
answered samples that are missed by both non-tool references, i.e., the
Tool-Free Agent and the Pure-Text Reasoner. This is the strictest
measure of tool-only solved cases. By contrast, \textit{Also solved
without tools} reports the proportion of tool-enabled successes that
are also solved by at least one non-tool setting, and is therefore the
complement of \textit{Tool-Enabled only}.

The last three columns further explain where this overlap comes from.
\textit{Also Tool-Free only} denotes samples solved by the Tool-Enabled
Agent and the Tool-Free Agent, but not by the Pure-Text Reasoner.
\textit{Also Pure-Text only} denotes samples solved by the Tool-Enabled
Agent and the Pure-Text Reasoner, but not by the Tool-Free Agent.
\textit{All three} denotes samples solved by all three settings. These
columns together decompose the non-tool overlap and show whether the
shared solved set comes from the inference-only tool-free variant, the
separately trained pure-text reference, or both.

The benchmark-level results make the main pattern in
Figure~\ref{fig:agent-success-coverage} more explicit. Across most benchmarks, the
Tool-Enabled Agent does not solve a largely separate set of examples.
Instead, most of its correct answers overlap with at least one non-tool
reference. This overlap is especially high on OCR and chart benchmarks,
where many tool-enabled successes are also reachable without executing
tools. Some benchmarks, such as TreeBench, MathVision, and LogicVista,
show larger tool-only portions, suggesting that tool access can still
help on localized subsets. However, these cases do not dominate the
overall picture. The table therefore supports the conclusion in
Section~\ref{sec:no-gains}: in the settings we study, tool access does
not consistently expand the solved set, and many correct tool-enabled
trajectories correspond to examples that are already solvable through
non-tool reasoning paths.

\section{Tool-Call Outcome Decomposition Details}
\label{tool-call-outcome-details}

\begin{table*}[p]
\centering
\scriptsize

\setlength{\heavyrulewidth}{0.08em}
\setlength{\lightrulewidth}{0.05em}
\setlength{\cmidrulewidth}{0.03em}

\caption{Full benchmark-level tool-call outcome decomposition for DeepEyesV2 and Thyme. TRR, TFR, TER, and TIR are computed on samples where the Tool-Enabled Agent calls tools, using either the Tool-Free Agent or the Pure-Text Reasoner as the non-tool baseline. Cell background colors follow the same blue colormap used in Table~\ref{tab:tool-call-benefit-task-avg}, with darker shades indicating larger values.}
\setlength{\tabcolsep}{3pt}
\renewcommand{\arraystretch}{1.12}
\setlength{\aboverulesep}{0pt}
\setlength{\belowrulesep}{0pt}
\resizebox{0.98\textwidth}{!}{%
\begin{tabular}{c>{\centering\arraybackslash}p{1.65cm}cccccccccc}
\toprule
\multirow{2}{*}{\textbf{Model}} & \multirow{2}{*}{\textbf{Task}} & \multirow{2}{*}{\textbf{Benchmark}} & \multicolumn{4}{c}{\textbf{vs Tool-Free Agent}} & \multicolumn{4}{c}{\textbf{vs Pure-Text Reasoner}} \\
\cmidrule(lr){4-7} \cmidrule(lr){8-11}
& & & \textbf{TRR} & \textbf{TFR} & \textbf{TER} & \textbf{TIR} & \textbf{TRR} & \textbf{TFR} & \textbf{TER} & \textbf{TIR} \\
\midrule
 \multirow{20}{*}{\textbf{DeepEyesV2}} & \multirow{5}{*}{\textbf{\shortstack{Real-World\\Understanding}}} & V* Bench & \cellcolor[HTML]{587896}\textcolor{white}{74.9} & \cellcolor[HTML]{C9D5E0}\textcolor{black}{13.1} & \cellcolor[HTML]{DCE3EB}\textcolor{black}{6.8} & \cellcolor[HTML]{DCE3EB}\textcolor{black}{5.2} & \cellcolor[HTML]{587896}\textcolor{white}{77.0} & \cellcolor[HTML]{C9D5E0}\textcolor{black}{12.6} & \cellcolor[HTML]{DCE3EB}\textcolor{black}{4.7} & \cellcolor[HTML]{DCE3EB}\textcolor{black}{5.8} \\
  &  & HRBench 4K & \cellcolor[HTML]{587896}\textcolor{white}{68.0} & \cellcolor[HTML]{B6C6D5}\textcolor{black}{17.5} & \cellcolor[HTML]{DCE3EB}\textcolor{black}{7.8} & \cellcolor[HTML]{DCE3EB}\textcolor{black}{6.8} & \cellcolor[HTML]{587896}\textcolor{white}{68.6} & \cellcolor[HTML]{B6C6D5}\textcolor{black}{16.1} & \cellcolor[HTML]{DCE3EB}\textcolor{black}{7.1} & \cellcolor[HTML]{DCE3EB}\textcolor{black}{8.1} \\
  &  & HRBench 8K & \cellcolor[HTML]{587896}\textcolor{white}{65.0} & \cellcolor[HTML]{B6C6D5}\textcolor{black}{23.4} & \cellcolor[HTML]{DCE3EB}\textcolor{black}{5.5} & \cellcolor[HTML]{DCE3EB}\textcolor{black}{6.1} & \cellcolor[HTML]{6E89A4}\textcolor{white}{62.9} & \cellcolor[HTML]{B6C6D5}\textcolor{black}{20.9} & \cellcolor[HTML]{DCE3EB}\textcolor{black}{7.6} & \cellcolor[HTML]{DCE3EB}\textcolor{black}{8.6} \\
  &  & Tree Bench & \cellcolor[HTML]{B6C6D5}\textcolor{black}{22.3} & \cellcolor[HTML]{83A0B7}\textcolor{black}{47.5} & \cellcolor[HTML]{B6C6D5}\textcolor{black}{19.3} & \cellcolor[HTML]{C9D5E0}\textcolor{black}{10.9} & \cellcolor[HTML]{B6C6D5}\textcolor{black}{28.7} & \cellcolor[HTML]{83A0B7}\textcolor{black}{44.6} & \cellcolor[HTML]{C9D5E0}\textcolor{black}{12.9} & \cellcolor[HTML]{C9D5E0}\textcolor{black}{13.9} \\
\cmidrule(lr){3-11}
  & & Avg & \cellcolor[HTML]{6E89A4}\textcolor{white}{59.1} & \cellcolor[HTML]{B6C6D5}\textcolor{black}{24.8} & \cellcolor[HTML]{DCE3EB}\textcolor{black}{9.0} & \cellcolor[HTML]{DCE3EB}\textcolor{black}{7.2} & \cellcolor[HTML]{6E89A4}\textcolor{white}{59.9} & \cellcolor[HTML]{B6C6D5}\textcolor{black}{22.8} & \cellcolor[HTML]{DCE3EB}\textcolor{black}{8.2} & \cellcolor[HTML]{DCE3EB}\textcolor{black}{9.2} \\
\cmidrule(lr){2-11}
  & \multirow{3}{*}{\textbf{OCR}} & OCR Bench & \cellcolor[HTML]{426986}\textcolor{white}{80.2} & \cellcolor[HTML]{C9D5E0}\textcolor{black}{12.0} & \cellcolor[HTML]{DCE3EB}\textcolor{black}{1.3} & \cellcolor[HTML]{DCE3EB}\textcolor{black}{6.5} & \cellcolor[HTML]{587896}\textcolor{white}{78.5} & \cellcolor[HTML]{DCE3EB}\textcolor{black}{8.7} & \cellcolor[HTML]{DCE3EB}\textcolor{black}{3.0} & \cellcolor[HTML]{DCE3EB}\textcolor{black}{9.8} \\
  &  & SEED 2 PLUS & \cellcolor[HTML]{587896}\textcolor{white}{65.1} & \cellcolor[HTML]{B6C6D5}\textcolor{black}{24.1} & \cellcolor[HTML]{DCE3EB}\textcolor{black}{4.7} & \cellcolor[HTML]{DCE3EB}\textcolor{black}{6.1} & \cellcolor[HTML]{6E89A4}\textcolor{white}{63.4} & \cellcolor[HTML]{B6C6D5}\textcolor{black}{23.7} & \cellcolor[HTML]{DCE3EB}\textcolor{black}{6.4} & \cellcolor[HTML]{DCE3EB}\textcolor{black}{6.5} \\
\cmidrule(lr){3-11}
  & & Avg & \cellcolor[HTML]{587896}\textcolor{white}{69.7} & \cellcolor[HTML]{B6C6D5}\textcolor{black}{20.4} & \cellcolor[HTML]{DCE3EB}\textcolor{black}{3.7} & \cellcolor[HTML]{DCE3EB}\textcolor{black}{6.2} & \cellcolor[HTML]{587896}\textcolor{white}{68.0} & \cellcolor[HTML]{B6C6D5}\textcolor{black}{19.1} & \cellcolor[HTML]{DCE3EB}\textcolor{black}{5.3} & \cellcolor[HTML]{DCE3EB}\textcolor{black}{7.5} \\
\cmidrule(lr){2-11}
  & \multirow{4}{*}{\textbf{Chart}} & CharXiv Descriptive & \cellcolor[HTML]{587896}\textcolor{white}{70.1} & \cellcolor[HTML]{B6C6D5}\textcolor{black}{17.4} & \cellcolor[HTML]{DCE3EB}\textcolor{black}{5.6} & \cellcolor[HTML]{DCE3EB}\textcolor{black}{6.9} & \cellcolor[HTML]{587896}\textcolor{white}{69.3} & \cellcolor[HTML]{B6C6D5}\textcolor{black}{15.3} & \cellcolor[HTML]{DCE3EB}\textcolor{black}{6.4} & \cellcolor[HTML]{DCE3EB}\textcolor{black}{9.0} \\
  &  & CharXiv Reasoning & \cellcolor[HTML]{99AEC3}\textcolor{black}{32.4} & \cellcolor[HTML]{83A0B7}\textcolor{black}{42.5} & \cellcolor[HTML]{C9D5E0}\textcolor{black}{11.3} & \cellcolor[HTML]{C9D5E0}\textcolor{black}{13.8} & \cellcolor[HTML]{B6C6D5}\textcolor{black}{30.7} & \cellcolor[HTML]{83A0B7}\textcolor{black}{42.0} & \cellcolor[HTML]{C9D5E0}\textcolor{black}{13.0} & \cellcolor[HTML]{B6C6D5}\textcolor{black}{14.3} \\
  &  & Chart QA & \cellcolor[HTML]{587896}\textcolor{white}{79.3} & \cellcolor[HTML]{DCE3EB}\textcolor{black}{8.1} & \cellcolor[HTML]{DCE3EB}\textcolor{black}{8.8} & \cellcolor[HTML]{DCE3EB}\textcolor{black}{3.8} & \cellcolor[HTML]{587896}\textcolor{white}{76.8} & \cellcolor[HTML]{DCE3EB}\textcolor{black}{7.7} & \cellcolor[HTML]{C9D5E0}\textcolor{black}{11.3} & \cellcolor[HTML]{DCE3EB}\textcolor{black}{4.2} \\
\cmidrule(lr){3-11}
  & & Avg & \cellcolor[HTML]{587896}\textcolor{white}{68.1} & \cellcolor[HTML]{B6C6D5}\textcolor{black}{17.6} & \cellcolor[HTML]{DCE3EB}\textcolor{black}{7.4} & \cellcolor[HTML]{DCE3EB}\textcolor{black}{6.8} & \cellcolor[HTML]{587896}\textcolor{white}{66.6} & \cellcolor[HTML]{B6C6D5}\textcolor{black}{16.3} & \cellcolor[HTML]{DCE3EB}\textcolor{black}{8.9} & \cellcolor[HTML]{DCE3EB}\textcolor{black}{8.1} \\
\cmidrule(lr){2-11}
  & \multirow{7}{*}{\textbf{Reasoning}} & MathVista & \cellcolor[HTML]{587896}\textcolor{white}{66.4} & \cellcolor[HTML]{B6C6D5}\textcolor{black}{20.1} & \cellcolor[HTML]{DCE3EB}\textcolor{black}{6.1} & \cellcolor[HTML]{DCE3EB}\textcolor{black}{7.4} & \cellcolor[HTML]{6E89A4}\textcolor{white}{64.2} & \cellcolor[HTML]{B6C6D5}\textcolor{black}{19.4} & \cellcolor[HTML]{DCE3EB}\textcolor{black}{8.3} & \cellcolor[HTML]{DCE3EB}\textcolor{black}{8.1} \\
  &  & MathVerse & \cellcolor[HTML]{99AEC3}\textcolor{black}{37.8} & \cellcolor[HTML]{99AEC3}\textcolor{black}{37.5} & \cellcolor[HTML]{C9D5E0}\textcolor{black}{13.9} & \cellcolor[HTML]{C9D5E0}\textcolor{black}{10.8} & \cellcolor[HTML]{99AEC3}\textcolor{black}{39.8} & \cellcolor[HTML]{99AEC3}\textcolor{black}{35.4} & \cellcolor[HTML]{C9D5E0}\textcolor{black}{12.0} & \cellcolor[HTML]{C9D5E0}\textcolor{black}{12.9} \\
  &  & MathVision & \cellcolor[HTML]{B6C6D5}\textcolor{black}{15.1} & \cellcolor[HTML]{6E89A4}\textcolor{white}{60.2} & \cellcolor[HTML]{C9D5E0}\textcolor{black}{11.3} & \cellcolor[HTML]{C9D5E0}\textcolor{black}{13.4} & \cellcolor[HTML]{B6C6D5}\textcolor{black}{15.1} & \cellcolor[HTML]{6E89A4}\textcolor{white}{60.0} & \cellcolor[HTML]{C9D5E0}\textcolor{black}{11.3} & \cellcolor[HTML]{C9D5E0}\textcolor{black}{13.6} \\
  &  & WeMath & \cellcolor[HTML]{83A0B7}\textcolor{black}{54.0} & \cellcolor[HTML]{B6C6D5}\textcolor{black}{23.6} & \cellcolor[HTML]{DCE3EB}\textcolor{black}{9.4} & \cellcolor[HTML]{C9D5E0}\textcolor{black}{13.0} & \cellcolor[HTML]{83A0B7}\textcolor{black}{54.0} & \cellcolor[HTML]{B6C6D5}\textcolor{black}{20.1} & \cellcolor[HTML]{DCE3EB}\textcolor{black}{9.4} & \cellcolor[HTML]{B6C6D5}\textcolor{black}{16.5} \\
  &  & DynaMath & \cellcolor[HTML]{83A0B7}\textcolor{black}{46.7} & \cellcolor[HTML]{B6C6D5}\textcolor{black}{31.2} & \cellcolor[HTML]{C9D5E0}\textcolor{black}{10.9} & \cellcolor[HTML]{C9D5E0}\textcolor{black}{11.2} & \cellcolor[HTML]{83A0B7}\textcolor{black}{46.7} & \cellcolor[HTML]{B6C6D5}\textcolor{black}{31.9} & \cellcolor[HTML]{C9D5E0}\textcolor{black}{10.9} & \cellcolor[HTML]{C9D5E0}\textcolor{black}{10.5} \\
  &  & LogicVista & \cellcolor[HTML]{99AEC3}\textcolor{black}{32.9} & \cellcolor[HTML]{99AEC3}\textcolor{black}{35.6} & \cellcolor[HTML]{B6C6D5}\textcolor{black}{15.2} & \cellcolor[HTML]{B6C6D5}\textcolor{black}{16.3} & \cellcolor[HTML]{B6C6D5}\textcolor{black}{30.9} & \cellcolor[HTML]{99AEC3}\textcolor{black}{38.9} & \cellcolor[HTML]{B6C6D5}\textcolor{black}{17.2} & \cellcolor[HTML]{C9D5E0}\textcolor{black}{13.0} \\
\cmidrule(lr){3-11}
  & & Avg & \cellcolor[HTML]{99AEC3}\textcolor{black}{39.8} & \cellcolor[HTML]{99AEC3}\textcolor{black}{37.1} & \cellcolor[HTML]{C9D5E0}\textcolor{black}{11.4} & \cellcolor[HTML]{C9D5E0}\textcolor{black}{11.6} & \cellcolor[HTML]{99AEC3}\textcolor{black}{40.1} & \cellcolor[HTML]{99AEC3}\textcolor{black}{36.5} & \cellcolor[HTML]{C9D5E0}\textcolor{black}{11.1} & \cellcolor[HTML]{C9D5E0}\textcolor{black}{12.3} \\
\cmidrule(lr){2-11}
  & \multicolumn{2}{c}{\textbf{Overall}} & \cellcolor[HTML]{83A0B7}\textcolor{black}{52.3} & \cellcolor[HTML]{B6C6D5}\textcolor{black}{29.0} & \cellcolor[HTML]{DCE3EB}\textcolor{black}{9.3} & \cellcolor[HTML]{DCE3EB}\textcolor{black}{9.4} & \cellcolor[HTML]{83A0B7}\textcolor{black}{52.0} & \cellcolor[HTML]{B6C6D5}\textcolor{black}{28.0} & \cellcolor[HTML]{DCE3EB}\textcolor{black}{9.6} & \cellcolor[HTML]{C9D5E0}\textcolor{black}{10.4} \\
\midrule
 \multirow{19}{*}{\textbf{Thyme}} & \multirow{5}{*}{\textbf{\shortstack{Real-World\\Understanding}}} & V* Bench & \cellcolor[HTML]{587896}\textcolor{white}{67.7} & \cellcolor[HTML]{B6C6D5}\textcolor{black}{16.1} & \cellcolor[HTML]{DCE3EB}\textcolor{black}{3.2} & \cellcolor[HTML]{C9D5E0}\textcolor{black}{12.9} & \cellcolor[HTML]{6E89A4}\textcolor{white}{64.5} & \cellcolor[HTML]{B6C6D5}\textcolor{black}{19.4} & \cellcolor[HTML]{DCE3EB}\textcolor{black}{6.5} & \cellcolor[HTML]{DCE3EB}\textcolor{black}{9.7} \\
  &  & HRBench 4K & \cellcolor[HTML]{587896}\textcolor{white}{73.0} & \cellcolor[HTML]{B6C6D5}\textcolor{black}{17.0} & \cellcolor[HTML]{DCE3EB}\textcolor{black}{2.0} & \cellcolor[HTML]{DCE3EB}\textcolor{black}{8.0} & \cellcolor[HTML]{587896}\textcolor{white}{67.0} & \cellcolor[HTML]{B6C6D5}\textcolor{black}{16.0} & \cellcolor[HTML]{DCE3EB}\textcolor{black}{8.0} & \cellcolor[HTML]{DCE3EB}\textcolor{black}{9.0} \\
  &  & HRBench 8K & \cellcolor[HTML]{587896}\textcolor{white}{74.3} & \cellcolor[HTML]{B6C6D5}\textcolor{black}{19.3} & \cellcolor[HTML]{DCE3EB}\textcolor{black}{3.7} & \cellcolor[HTML]{DCE3EB}\textcolor{black}{2.8} & \cellcolor[HTML]{587896}\textcolor{white}{71.6} & \cellcolor[HTML]{B6C6D5}\textcolor{black}{19.3} & \cellcolor[HTML]{DCE3EB}\textcolor{black}{6.4} & \cellcolor[HTML]{DCE3EB}\textcolor{black}{2.8} \\
  &  & Tree Bench & \cellcolor[HTML]{99AEC3}\textcolor{black}{40.0} & \cellcolor[HTML]{83A0B7}\textcolor{black}{53.3} & \cellcolor[HTML]{EFF2F6}\textcolor{black}{0.0} & \cellcolor[HTML]{DCE3EB}\textcolor{black}{6.7} & \cellcolor[HTML]{99AEC3}\textcolor{black}{33.3} & \cellcolor[HTML]{99AEC3}\textcolor{black}{33.3} & \cellcolor[HTML]{DCE3EB}\textcolor{black}{6.7} & \cellcolor[HTML]{B6C6D5}\textcolor{black}{26.7} \\
\cmidrule(lr){3-11}
  & & Avg & \cellcolor[HTML]{587896}\textcolor{white}{71.0} & \cellcolor[HTML]{B6C6D5}\textcolor{black}{20.0} & \cellcolor[HTML]{DCE3EB}\textcolor{black}{2.7} & \cellcolor[HTML]{DCE3EB}\textcolor{black}{6.3} & \cellcolor[HTML]{587896}\textcolor{white}{66.7} & \cellcolor[HTML]{B6C6D5}\textcolor{black}{18.8} & \cellcolor[HTML]{DCE3EB}\textcolor{black}{7.1} & \cellcolor[HTML]{DCE3EB}\textcolor{black}{7.5} \\
\cmidrule(lr){2-11}
  & \multirow{3}{*}{\textbf{OCR}} & OCR Bench & \cellcolor[HTML]{426986}\textcolor{white}{85.7} & \cellcolor[HTML]{C9D5E0}\textcolor{black}{11.8} & \cellcolor[HTML]{EFF2F6}\textcolor{black}{0.3} & \cellcolor[HTML]{DCE3EB}\textcolor{black}{2.2} & \cellcolor[HTML]{426986}\textcolor{white}{82.9} & \cellcolor[HTML]{DCE3EB}\textcolor{black}{8.0} & \cellcolor[HTML]{DCE3EB}\textcolor{black}{3.0} & \cellcolor[HTML]{DCE3EB}\textcolor{black}{6.1} \\
  &  & SEED 2 PLUS & \cellcolor[HTML]{587896}\textcolor{white}{68.1} & \cellcolor[HTML]{B6C6D5}\textcolor{black}{23.4} & \cellcolor[HTML]{DCE3EB}\textcolor{black}{2.1} & \cellcolor[HTML]{DCE3EB}\textcolor{black}{6.4} & \cellcolor[HTML]{587896}\textcolor{white}{66.0} & \cellcolor[HTML]{B6C6D5}\textcolor{black}{25.5} & \cellcolor[HTML]{DCE3EB}\textcolor{black}{4.3} & \cellcolor[HTML]{DCE3EB}\textcolor{black}{4.3} \\
\cmidrule(lr){3-11}
  & & Avg & \cellcolor[HTML]{426986}\textcolor{white}{83.7} & \cellcolor[HTML]{C9D5E0}\textcolor{black}{13.2} & \cellcolor[HTML]{EFF2F6}\textcolor{black}{0.5} & \cellcolor[HTML]{DCE3EB}\textcolor{black}{2.7} & \cellcolor[HTML]{426986}\textcolor{white}{81.0} & \cellcolor[HTML]{C9D5E0}\textcolor{black}{10.0} & \cellcolor[HTML]{DCE3EB}\textcolor{black}{3.2} & \cellcolor[HTML]{DCE3EB}\textcolor{black}{5.9} \\
\cmidrule(lr){2-11}
  & \multirow{4}{*}{\textbf{Chart}} & CharXiv Descriptive & \cellcolor[HTML]{99AEC3}\textcolor{black}{40.0} & \cellcolor[HTML]{99AEC3}\textcolor{black}{40.0} & \cellcolor[HTML]{EFF2F6}\textcolor{black}{0.0} & \cellcolor[HTML]{B6C6D5}\textcolor{black}{20.0} & \cellcolor[HTML]{99AEC3}\textcolor{black}{40.0} & \cellcolor[HTML]{B6C6D5}\textcolor{black}{20.0} & \cellcolor[HTML]{EFF2F6}\textcolor{black}{0.0} & \cellcolor[HTML]{99AEC3}\textcolor{black}{40.0} \\
  &  & CharXiv Reasoning & \cellcolor[HTML]{EFF2F6}\textcolor{black}{0.0} & \cellcolor[HTML]{426986}\textcolor{white}{100.0} & \cellcolor[HTML]{EFF2F6}\textcolor{black}{0.0} & \cellcolor[HTML]{EFF2F6}\textcolor{black}{0.0} & \cellcolor[HTML]{EFF2F6}\textcolor{black}{0.0} & \cellcolor[HTML]{83A0B7}\textcolor{black}{50.0} & \cellcolor[HTML]{EFF2F6}\textcolor{black}{0.0} & \cellcolor[HTML]{83A0B7}\textcolor{black}{50.0} \\
  &  & Chart QA & \cellcolor[HTML]{426986}\textcolor{white}{88.9} & \cellcolor[HTML]{C9D5E0}\textcolor{black}{11.1} & \cellcolor[HTML]{EFF2F6}\textcolor{black}{0.0} & \cellcolor[HTML]{EFF2F6}\textcolor{black}{0.0} & \cellcolor[HTML]{83A0B7}\textcolor{black}{44.4} & \cellcolor[HTML]{EFF2F6}\textcolor{black}{0.0} & \cellcolor[HTML]{83A0B7}\textcolor{black}{44.4} & \cellcolor[HTML]{C9D5E0}\textcolor{black}{11.1} \\
\cmidrule(lr){3-11}
  & & Avg & \cellcolor[HTML]{6E89A4}\textcolor{white}{62.5} & \cellcolor[HTML]{B6C6D5}\textcolor{black}{31.2} & \cellcolor[HTML]{EFF2F6}\textcolor{black}{0.0} & \cellcolor[HTML]{DCE3EB}\textcolor{black}{6.2} & \cellcolor[HTML]{99AEC3}\textcolor{black}{37.5} & \cellcolor[HTML]{C9D5E0}\textcolor{black}{12.5} & \cellcolor[HTML]{B6C6D5}\textcolor{black}{25.0} & \cellcolor[HTML]{B6C6D5}\textcolor{black}{25.0} \\
\cmidrule(lr){2-11}
  & \multirow{6}{*}{\textbf{Reasoning}} & MathVista & \cellcolor[HTML]{EFF2F6}\textcolor{black}{0.0} & \cellcolor[HTML]{426986}\textcolor{white}{100.0} & \cellcolor[HTML]{EFF2F6}\textcolor{black}{0.0} & \cellcolor[HTML]{EFF2F6}\textcolor{black}{0.0} & \cellcolor[HTML]{EFF2F6}\textcolor{black}{0.0} & \cellcolor[HTML]{426986}\textcolor{white}{100.0} & \cellcolor[HTML]{EFF2F6}\textcolor{black}{0.0} & \cellcolor[HTML]{EFF2F6}\textcolor{black}{0.0} \\
  &  & MathVerse & \cellcolor[HTML]{99AEC3}\textcolor{black}{37.5} & \cellcolor[HTML]{99AEC3}\textcolor{black}{39.6} & \cellcolor[HTML]{B6C6D5}\textcolor{black}{22.9} & \cellcolor[HTML]{EFF2F6}\textcolor{black}{0.0} & \cellcolor[HTML]{99AEC3}\textcolor{black}{39.6} & \cellcolor[HTML]{99AEC3}\textcolor{black}{35.4} & \cellcolor[HTML]{B6C6D5}\textcolor{black}{20.8} & \cellcolor[HTML]{DCE3EB}\textcolor{black}{4.2} \\
  &  & MathVision & \cellcolor[HTML]{C9D5E0}\textcolor{black}{12.5} & \cellcolor[HTML]{587896}\textcolor{white}{75.0} & \cellcolor[HTML]{EFF2F6}\textcolor{black}{0.0} & \cellcolor[HTML]{C9D5E0}\textcolor{black}{12.5} & \cellcolor[HTML]{C9D5E0}\textcolor{black}{12.5} & \cellcolor[HTML]{6E89A4}\textcolor{white}{62.5} & \cellcolor[HTML]{EFF2F6}\textcolor{black}{0.0} & \cellcolor[HTML]{B6C6D5}\textcolor{black}{25.0} \\
  &  & WeMath & \cellcolor[HTML]{83A0B7}\textcolor{black}{50.0} & \cellcolor[HTML]{83A0B7}\textcolor{black}{50.0} & \cellcolor[HTML]{EFF2F6}\textcolor{black}{0.0} & \cellcolor[HTML]{EFF2F6}\textcolor{black}{0.0} & \cellcolor[HTML]{EFF2F6}\textcolor{black}{0.0} & \cellcolor[HTML]{EFF2F6}\textcolor{black}{0.0} & \cellcolor[HTML]{83A0B7}\textcolor{black}{50.0} & \cellcolor[HTML]{83A0B7}\textcolor{black}{50.0} \\
  &  & DynaMath & \cellcolor[HTML]{B6C6D5}\textcolor{black}{29.7} & \cellcolor[HTML]{99AEC3}\textcolor{black}{40.6} & \cellcolor[HTML]{B6C6D5}\textcolor{black}{25.0} & \cellcolor[HTML]{DCE3EB}\textcolor{black}{4.7} & \cellcolor[HTML]{B6C6D5}\textcolor{black}{28.1} & \cellcolor[HTML]{99AEC3}\textcolor{black}{35.9} & \cellcolor[HTML]{B6C6D5}\textcolor{black}{26.6} & \cellcolor[HTML]{DCE3EB}\textcolor{black}{9.4} \\
\cmidrule(lr){3-11}
  & & Avg & \cellcolor[HTML]{B6C6D5}\textcolor{black}{31.7} & \cellcolor[HTML]{83A0B7}\textcolor{black}{43.1} & \cellcolor[HTML]{B6C6D5}\textcolor{black}{22.0} & \cellcolor[HTML]{DCE3EB}\textcolor{black}{3.3} & \cellcolor[HTML]{B6C6D5}\textcolor{black}{30.9} & \cellcolor[HTML]{99AEC3}\textcolor{black}{37.4} & \cellcolor[HTML]{B6C6D5}\textcolor{black}{22.8} & \cellcolor[HTML]{DCE3EB}\textcolor{black}{8.9} \\
\cmidrule(lr){2-11}
  & \multicolumn{2}{c}{\textbf{Overall}} & \cellcolor[HTML]{587896}\textcolor{white}{71.3} & \cellcolor[HTML]{B6C6D5}\textcolor{black}{20.3} & \cellcolor[HTML]{DCE3EB}\textcolor{black}{4.5} & \cellcolor[HTML]{DCE3EB}\textcolor{black}{4.0} & \cellcolor[HTML]{587896}\textcolor{white}{67.9} & \cellcolor[HTML]{B6C6D5}\textcolor{black}{17.0} & \cellcolor[HTML]{DCE3EB}\textcolor{black}{7.8} & \cellcolor[HTML]{DCE3EB}\textcolor{black}{7.2} \\
\bottomrule
\end{tabular}%
}
\label{tab:app-tool-call-benefit-full}
\end{table*}

Table~\ref{tab:app-tool-call-benefit-full} provides the benchmark-level
decomposition behind the task-family averages reported in
Table~\ref{tab:tool-call-benefit-task-avg}. The table keeps the same four-way
outcome taxonomy as the main text, but reports it separately for each
benchmark and for both non-tool references. All rates are computed on
the subset of samples where the Tool-Enabled Agent actually invokes a
tool, so the table focuses specifically on trajectories with observed
tool-use behavior rather than on all evaluated samples.

This finer-grained view shows that the task-family trends in the main
text are not driven by a single benchmark. On OCR and chart-related
benchmarks, tool-called samples frequently fall into shared-correctness
regions with the non-tool references, indicating that many successful
tool-use trajectories are not unique to tool-enabled inference. On
reasoning benchmarks, the distribution shifts more strongly toward
shared failures, especially on harder datasets such as MathVision,
MathVerse, and LogicVista. These cases show that tool invocation often
continues to occur on difficult samples, but does not necessarily
convert them into correct answers.

The benchmark-level results also reveal where tool use is more
locally beneficial. Some benchmarks contain non-negligible effective
correction rates, suggesting that tool calls can help on particular
subsets. However, these gains are not the dominant pattern across the
full benchmark suite. Together with the solved-set coverage in
Table~\ref{tab:app-solved-set-benchmark}, this table provides a more detailed
account of why the aggregate tool-enabled performance should not be
interpreted as direct evidence of broad capability expansion.

\section{Process-Level Attribution Details}
\label{app:process-attribution-details}

To complement the correctness-based decompositions, we further annotate
the role played by each tool call inside the reasoning trajectory. This
analysis is conducted on the full tool-enabled outputs of DeepEyesV2 and
Thyme across all 15 benchmarks. We parse each model output, identify
whether the agent actually invokes a tool, and keep only trajectories
with observed tool-use behavior. The annotation therefore focuses on
what happens when the tool-use mechanism is explicitly activated, rather
than on all evaluated samples.

We use Qwen3-VL-30B as the judge model for process-level attribution.
For each tool-using trajectory, the judge is given the question, the
model's reasoning context, the tool-call content, the returned tool
output, and the final answer. The full judge prompt is provided below.
The prompt is designed to avoid judging a tool call only by the final
answer. In particular, it explicitly instructs the judge not to mark a
tool output as useful merely because the final prediction is correct.
For example, a tool call that only re-displays the original image, such
as a simple \texttt{plt.imshow(image\_1)} operation without cropping,
zooming, OCR, enhancement, measurement, or parsing, should not be
treated as providing novel information. Such cases are usually
confirmatory or non-contributory rather than genuine evidence of
tool-enabled reasoning.

The judge assigns labels along three strictly separated dimensions:
\textit{Information Gain}, \textit{Tool Output Quality}, and
\textit{Integration Status}. For \textit{Information Gain}, the allowed
labels are \textit{Novel}, \textit{Confirmatory}, and
\textit{Irrelevant}. For \textit{Tool Output Quality}, the allowed
labels are \textit{Useful/Correct}, \textit{Partially useful}, and
\textit{Wrong/Failed}. For \textit{Integration Status}, the allowed
labels are \textit{Used correctly}, \textit{Ignored}, and
\textit{Misused/Misinterpreted}. During post-processing, we normalize
\textit{Useful/Correct} to \textit{Correct}, and \textit{Partially
useful} to \textit{Partially correct}. The prompt also emphasizes that
\textit{Novel} should be assigned conservatively, since many tool calls
only reformat, restate, or visually redisplay information already
available to the model.

We then map the three-dimensional labels into mutually exclusive
trajectory roles using a priority-based rule. If the integration label
is \textit{Misused / Misinterpreted}, the trajectory is assigned to
\textit{Misuse / other}, regardless of the information
gain or output quality labels. A trajectory is assigned to
\textit{Genuine contribution} only when the tool output is
\textit{Novel}, \textit{Correct}, and \textit{Used correctly}. If the
tool output is \textit{Novel} and \textit{Correct} or
\textit{Partially correct}, but the model ignores it, the trajectory is
assigned to \textit{Missed opportunity}. If the tool output is
\textit{Confirmatory}, has \textit{Correct} or \textit{Partially
correct} quality, and is either used or ignored, the trajectory is
assigned to \textit{Redundant confirmation}. If the tool output is
\textit{Wrong / Failed} and is either ignored or used, the trajectory is
assigned to \textit{Failed or non-contributory call}. Remaining rare
label combinations are assigned to \textit{Other}.

For visualization, we merge these fine-grained roles into four display
categories. \textit{Redundant confirmation} and \textit{Genuine
contribution} are kept as separate categories. \textit{Failed or
non-contributory call} is displayed as \textit{Failed /
non-contributory}. The remaining roles, including \textit{Misuse or
harmful integration}, \textit{Missed opportunity}, and \textit{Other},
are grouped as \textit{Misuse / other}. Figure~\ref{fig:tool-process-analysis}
shows the resulting task-family-level role distributions for DeepEyesV2
and Thyme.
    
Tables~\ref{tab:appendix-tool-process-deepeyesv2} and
\ref{tab:appendix-tool-process-thyme} provide the corresponding
benchmark-level breakdowns. These tables expand Figure~\ref{fig:tool-process-analysis}
by reporting the role distribution for each individual benchmark rather
than only for each task family. They make it possible to see whether a
task-family trend is broadly shared across benchmarks or driven by a
small number of datasets. For example, they show which OCR or chart
benchmarks contribute most to redundant confirmation, and which
reasoning benchmarks contribute more to failed or non-contributory
tool calls. Thus, the tables serve as the fine-grained audit trail
behind the process-level summary in the main text.

The first two subfigures in Figure~\ref{fig:tool-process-analysis} show
100\%-stacked role distributions over four task families: real-world
understanding, OCR, chart understanding, and mathematical reasoning.
The third subfigure further examines the subset labeled as
\textit{Genuine tool contribution}. For this subset, we check whether
the same agent can answer the sample correctly when tool use is disabled
at inference time. This tests whether a trajectory that appears
genuinely tool-contributive at the process level also corresponds to an
example that the agent could not solve without tools.

The results show that even the process-level ``genuine'' subset contains
substantial overlap with tool-free correctness. Among trajectories
labeled as \textit{Genuine contribution}, 59.7\% of DeepEyesV2
samples and 79.4\% of Thyme samples are still answered correctly by the
corresponding Tool-Free Agent. Conversely, only 40.3\% and 20.6\% of
these subsets, respectively, correspond to samples that the Tool-Free
Agent gets wrong. This indicates that even when the tool output is
judged to be novel, correct, and integrated, the tool call often does
not uniquely expand the agent's solvable set.

To assess the reliability of the judge-based attribution, we further conduct a manual audit on a random 40\% subset of the judged tool-using trajectories. Human inspection shows broad agreement with the judge labels on the major role categories, especially the distinctions among redundant confirmation, genuine contribution, and failed or non-contributory calls. Most disagreements occur in borderline cases where the tool output is partially useful, or where the trajectory makes it difficult to determine whether the final answer actually depends on the tool output. No systematic disagreement was found that would materially alter the dominant role trends observed in Figure~\ref{fig:tool-process-analysis} and Tables~\ref{tab:appendix-tool-process-deepeyesv2} and~\ref{tab:appendix-tool-process-thyme}. We therefore use the judge annotations as a process-level diagnostic signal rather than as ground-truth causal attribution.


\FloatBarrier
\paragraph{Judge prompt for tool-use process attribution.}
The following prompt is used to annotate tool-called trajectories in the tool-use process analysis.

\begin{Verbatim}[breaklines=true,breakanywhere=true,fontsize=\small]
You are an expert judge for multimodal agent tool-use analysis.

Your task is to inspect an agent trajectory and label the role of tool use using three dimensions.

Judge only from observable evidence in the trajectory. Do NOT give credit to tool use just because:
- the final answer is correct,
- the model says things became "clearer",
- or a tool call happened.

You must distinguish between truly new evidence and mere repetition / self-confirmation.
You must also keep the three dimensions strictly separate. Never reuse a label from one dimension in another dimension.

Definitions:
1. Information Gain
- Novel: the tool provides new task-relevant information not already available from the model's unaided reasoning.
- Confirmatory: the tool mainly confirms or repeats an already available conclusion.
- Irrelevant: the tool output is not materially useful for solving the task.

2. Tool Output Quality
- Useful/Correct: the tool output is correct and meaningfully usable.
- Partially useful: the tool output is partially correct, incomplete, or only weakly useful.
- Wrong/Failed: the tool output is wrong, failed, or unusable.

3. Integration Status
- Used correctly: the final reasoning meaningfully uses the tool output in a correct way.
- Ignored: useful or potentially useful tool output is not actually used.
- Misused/Misinterpreted: the model uses the tool output incorrectly or is misled by it.

Important decision rules:
1. Novel should be rare. Only assign Novel if the tool output clearly reveals or computes information
   that was not already available from the pre-tool reasoning.
2. If the tool only redisplays the original image without crop, zoom, OCR, enhancement, measurement,
   parsing, or other transformation, do NOT label it as Novel.
3. If the tool output is empty, trivial, or only repeats the original image view, it should usually be
   Confirmatory or Irrelevant, not Novel.
4. Do not infer tool usefulness from final correctness. A correct final answer can still correspond to
   Confirmatory, Partially useful, or even Irrelevant tool use.
5. Useful/Correct requires that the tool output itself is materially informative and correct. If it adds
   little or no new evidence, prefer Partially useful or Wrong/Failed as appropriate.
6. If the post-tool answer mostly repeats the same conclusion already stated before the tool call, that is
   evidence against Novel information gain.
7. Irrelevant is ONLY valid for information_gain. It is NOT a valid value for tool_output_quality or
   integration_status.
8. If the tool output is empty, blank, trivial, or unusable, then tool_output_quality should usually be
   Wrong/Failed.
9. If the tool output is merely the same unprocessed image view, with no crop / zoom / OCR / enhancement /
   measurement / parsing, then tool_output_quality should usually be Partially useful at most, and often
   Wrong/Failed if the execution adds nothing.
10. If the tool output adds nothing and the final answer comes from the same pre-tool reasoning, prefer
    integration_status = Ignored, not Used correctly.

Field-specific allowed values:
- information_gain must be exactly one of: Novel, Confirmatory, Irrelevant
- tool_output_quality must be exactly one of: Useful/Correct, Partially useful, Wrong/Failed
- integration_status must be exactly one of: Used correctly, Ignored, Misused/Misinterpreted

Return ONLY a JSON object with exactly these keys:
{
  "information_gain": "...",
  "tool_output_quality": "...",
  "integration_status": "...",
  "confidence": 0-100,
  "rationale": "short explanation"
}

User prompt template:
Analyze the following multimodal agent trajectory.

Metadata:
- Model setting: {model_key}
- Benchmark: {benchmark}
- Sample index: {sample_index}
- Tool use times: {tool_use_times}
- Final score: {score}

Trajectory:
<trajectory>
{trajectory_text}
</trajectory>

Return only the JSON object.
\end{Verbatim}

\FloatBarrier

\begin{table*}[t]
\centering
\small
\caption{Benchmark-level tool-use role distribution for DeepEyesV2. Percentages are computed over samples where the tool-enabled agent calls tools within each benchmark. Entries marked ``--'' indicate that the agent did not call tools on that benchmark.}
\label{tab:appendix-tool-process-deepeyesv2}
\resizebox{\textwidth}{!}{%
\begin{tabular}{llrrrrr}
\toprule
Task & Benchmark & $N$ & Redundant confirmation & Genuine contribution & Failed / non-contributory & Misuse / other \\
\midrule
RWU & V* & 191 & 23.6 & 11.5 & 52.9 & 12.0 \\
RWU & HRBench-4K & 800 & 24.8 & 7.9 & 59.9 & 7.5 \\
RWU & HRBench-8K & 800 & 22.5 & 7.9 & 62.3 & 7.4 \\
RWU & TreeBench & 404 & 13.9 & 3.2 & 79.2 & 3.7 \\
\midrule
OCR & OCRBench & 999 & 54.3 & 3.8 & 40.8 & 1.1 \\
OCR & SEED2 PLUS & 2276 & 67.7 & 7.0 & 22.4 & 3.0 \\
\midrule
Chart & CharXiv-Desc & 3999 & 44.8 & 16.7 & 37.6 & 0.9 \\
Chart & CharXiv-Reason & 1000 & 48.9 & 11.5 & 36.7 & 2.9 \\
Chart & ChartQA & 2500 & 78.7 & 15.2 & 5.1 & 1.0 \\
\midrule
Reasoning & MathVista & 1000 & 57.5 & 26.9 & 12.6 & 3.0 \\
Reasoning & MathVerse & 3940 & 43.9 & 50.3 & 0.6 & 5.2 \\
Reasoning & MathVision & 3039 & 54.3 & 31.1 & 8.2 & 6.4 \\
Reasoning & WeMath & 1740 & 50.0 & 44.4 & 3.6 & 2.0 \\
Reasoning & DynaMath & 5009 & 42.4 & 41.3 & 13.4 & 2.9 \\
Reasoning & LogicVista & 447 & 63.5 & 21.0 & 13.0 & 2.5 \\
\bottomrule
\end{tabular}%
}
\end{table*}

\begin{table*}[t]
\centering
\small
\caption{Benchmark-level tool-use role distribution for Thyme. Percentages are computed over samples where the tool-enabled agent calls tools within each benchmark. Entries marked ``--'' indicate that the agent did not call tools on that benchmark.}
\label{tab:appendix-tool-process-thyme}
\resizebox{\textwidth}{!}{%
\begin{tabular}{llrrrrr}
\toprule
Task & Benchmark & $N$ & Redundant confirmation & Genuine contribution & Failed / non-contributory & Misuse / other \\
\midrule
RWU & V* & 31 & 0.0 & 100.0 & 0.0 & 0.0 \\
RWU & HRBench-4K & 100 & 0.0 & 100.0 & 0.0 & 0.0 \\
RWU & HRBench-8K & 109 & 0.0 & 93.6 & 5.5 & 0.9 \\
RWU & TreeBench & 15 & 0.0 & 86.7 & 6.7 & 6.7 \\
\midrule
OCR & OCRBench & 363 & 0.0 & 95.0 & 3.3 & 1.7 \\
OCR & SEED2 PLUS & 47 & 0.0 & 100.0 & 0.0 & 0.0 \\
\midrule
Chart & CharXiv-Desc & 5 & 0.0 & 100.0 & 0.0 & 0.0 \\
Chart & CharXiv-Reason & 2 & 0.0 & 50.0 & 0.0 & 50.0 \\
Chart & ChartQA & 9 & 100.0 & 0.0 & 0.0 & 0.0 \\
\midrule
Reasoning & MathVista & 1 & 100.0 & 0.0 & 0.0 & 0.0 \\
Reasoning & MathVerse & 48 & 25.0 & 75.0 & 0.0 & 0.0 \\
Reasoning & MathVision & 8 & 37.5 & 62.5 & 0.0 & 0.0 \\
Reasoning & WeMath & 2 & 100.0 & 0.0 & 0.0 & 0.0 \\
Reasoning & DynaMath & 64 & 15.6 & 73.4 & 0.0 & 10.9 \\
Reasoning & LogicVista & 0 & -- & -- & -- & -- \\
\bottomrule
\end{tabular}%
}
\end{table*}

\section{Generated-Token Statistics}
\label{app:token-statistics}

In addition to accuracy and solved-set coverage, we also examine whether
tool-enabled inference brings a compensating efficiency benefit. Even if
tool access does not consistently improve final-answer correctness, it
might still be useful if external tool execution reduces the amount of
model generation required to reach an answer. The main text summarizes
this analysis at the task-family level; here we provide the detailed
token statistics.

\begin{figure*}[!t]
  \centering
  \begin{subfigure}{0.49\textwidth}
    \centering
    \includegraphics[width=\textwidth]{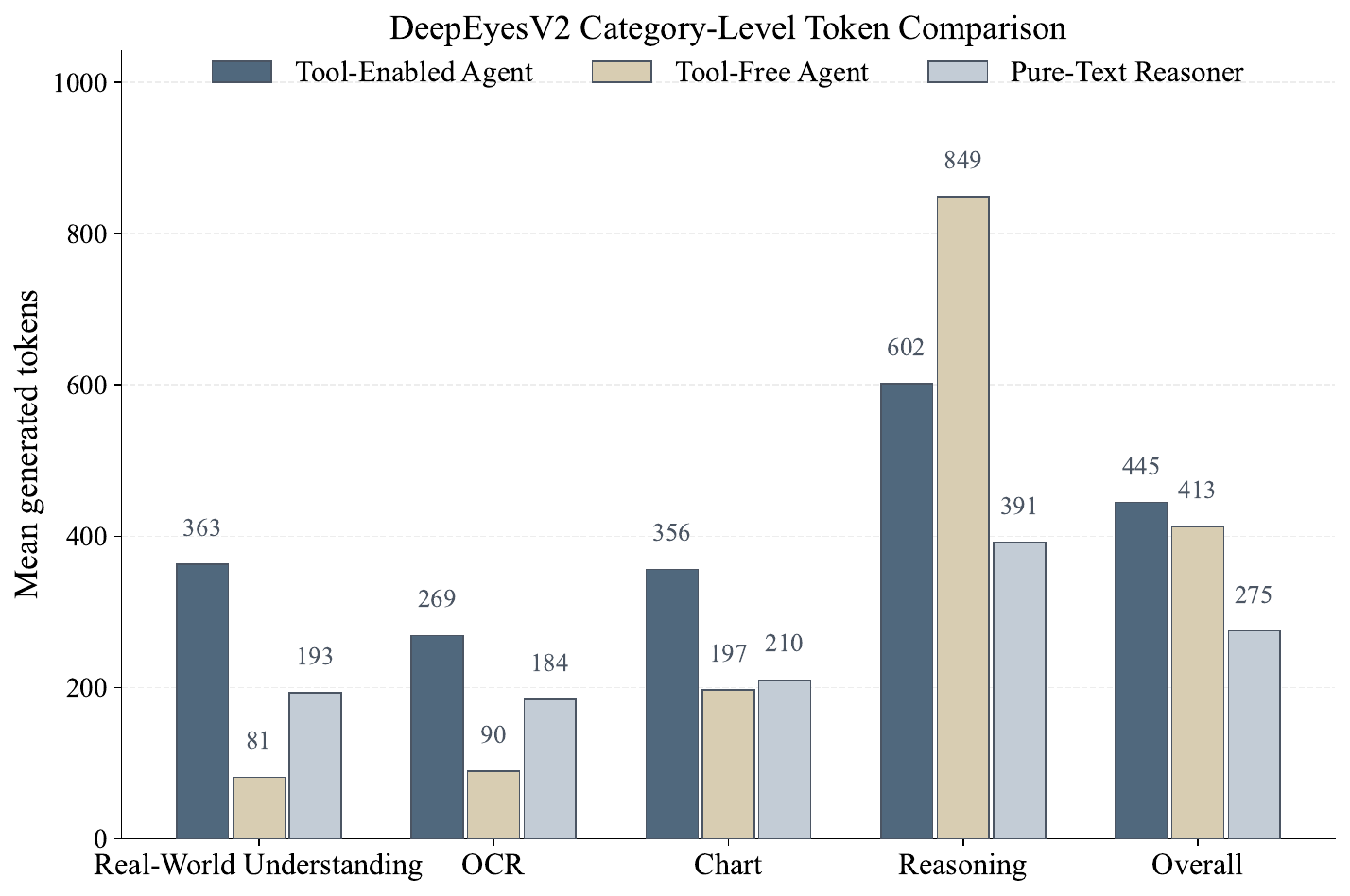}
    \caption{DeepEyesV2 token totals comparison.}
    \label{fig:token-deepeyesv2}
  \end{subfigure}
  \hfill
  \begin{subfigure}{0.49\textwidth}
    \centering
    \includegraphics[width=\textwidth]{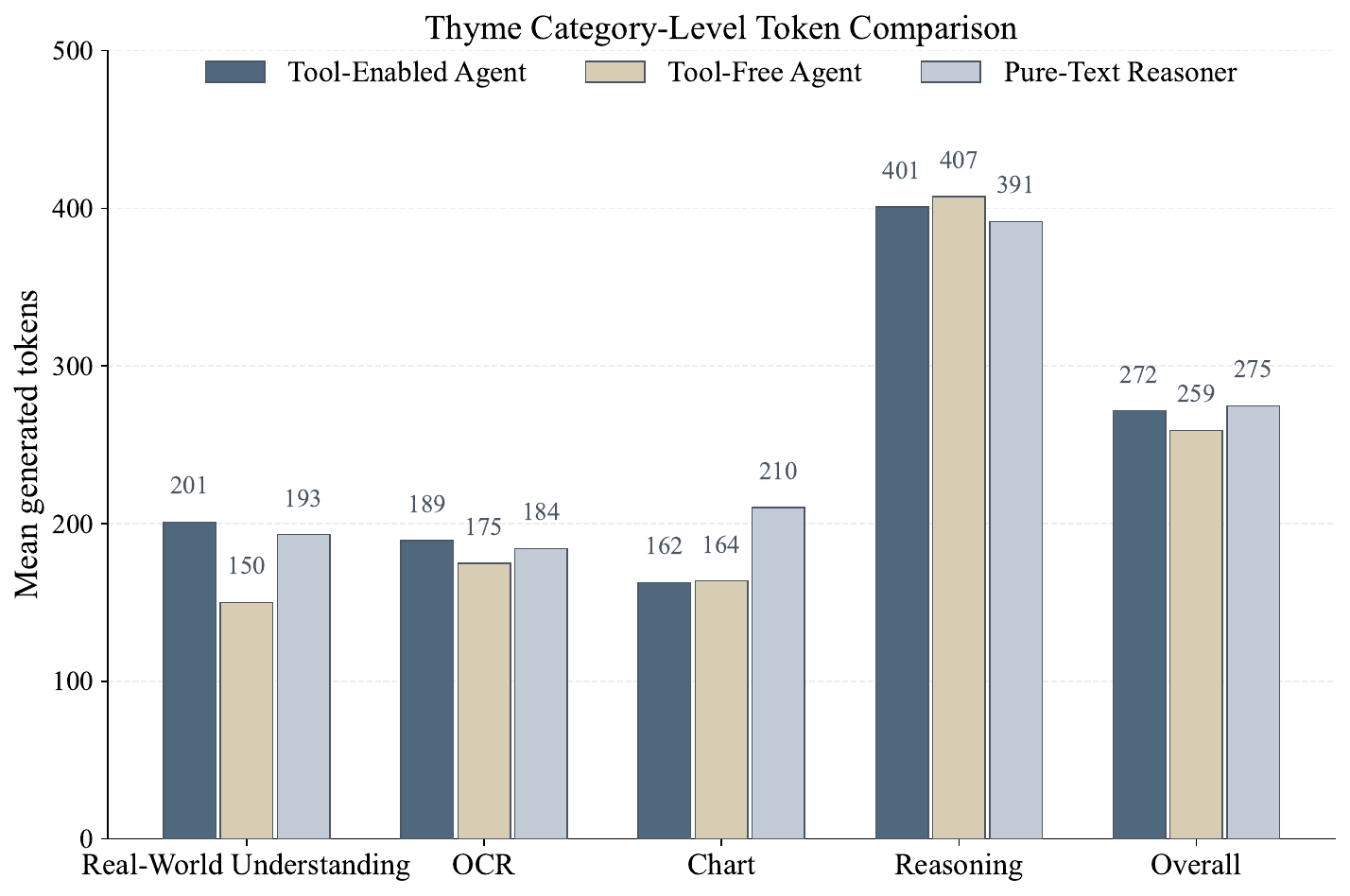}
    \caption{Thyme token totals comparison.}
    \label{fig:token-thyme}
  \end{subfigure}
  \caption{Average generated-token counts across task families for the Tool-Enabled Agent, the Tool-Free Agent, and the Pure-Text Reasoner.}
  \label{fig:token-totals-comparison}
\end{figure*}

Figure~\ref{fig:token-totals-comparison} shows the average number of
generated tokens across task families for the Tool-Enabled Agent, the
Tool-Free Agent, and the Pure-Text Reasoner. Table~\ref{tab:app-token-cost-benchmark}
further reports the benchmark-level token counts. We count tokens
generated by the model, including reasoning text and tool-call messages.
External tool outputs are not treated as model-generated tokens, since
they are produced by the execution environment rather than by the agent
itself.

The results show that tool-enabled inference does not reliably reduce
generation cost. For DeepEyesV2, the Tool-Enabled Agent often produces
more tokens than both non-tool settings, and its overall average remains
higher than the Tool-Free Agent and the Pure-Text Reasoner. Thyme shows
smaller differences, but the same qualitative pattern holds: enabling
tools does not lead to a consistent reduction in generated tokens across
benchmarks. In several task families, tool-enabled inference introduces
additional reasoning and tool-call overhead without a corresponding
decrease in final generation length.

The benchmark-level results in Table~\ref{tab:app-token-cost-benchmark}
make this pattern more explicit. Token counts vary across datasets and
agents, but the tool-enabled setting is not systematically more
efficient. This complements the accuracy and solved-set analyses: in
the settings we study, tool access neither consistently expands the set
of solved examples nor provides a reliable token-saving benefit.

\begin{table*}[t]
\centering
\small
\setlength{\tabcolsep}{4pt}
\renewcommand{\arraystretch}{1.08}
\caption{Benchmark-level token cost comparison. Values report the average number of generated tokens per sample. Score gap is computed as the Tool-Enabled Agent score minus the Pure-Text Reasoner score on the same benchmark.}
\label{tab:app-token-cost-benchmark}
\resizebox{\textwidth}{!}{%
\begin{tabular}{llrrrr}
\toprule
Model & Benchmark & Pure-Text Reasoner & Tool-Free Agent & Tool-Enabled Agent & Score gap \\
\midrule
DeepEyesV2 & V* Bench & 141.4 & 55.6 & 354.4 & -1.1 \\
 & HRBench 4K & 215.3 & 66.0 & 324.2 & -1.1 \\
 & HRBench 8K & 209.8 & 69.5 & 340.9 & -0.3 \\
 & Tree Bench & 205.3 & 134.5 & 433.3 & -0.7 \\
 & OCR Bench & 163.6 & 67.3 & 212.8 & -2.4 \\
 & SEED 2 PLUS & 204.5 & 111.7 & 324.9 & -0.1 \\
 & CharXiv descriptive & 208.0 & 167.3 & 325.9 & 3.2 \\
 & CharXiv reasoning & 243.7 & 305.1 & 481.3 & -1.3 \\
 & Chart QA & 178.8 & 118.9 & 262.1 & 8.8 \\
 & MathVista & 277.1 & 378.1 & 444.0 & 0.2 \\
 & MathVerse & 404.1 & 916.8 & 572.8 & -0.3 \\
 & MathVision & 541.1 & 1626.0 & 904.0 & -1.8 \\
 & WeMath & 371.8 & 773.7 & 483.0 & -7.1 \\
 & DynaMath & 374.1 & 644.2 & 588.1 & 3.1 \\
 & LogicVista & 380.6 & 753.7 & 617.3 & 4.3 \\
\midrule
Thyme & V* Bench & 141.4 & 140.4 & 214.8 & 0.0 \\
 & HRBench 4K & 215.3 & 147.8 & 197.3 & 0.5 \\
 & HRBench 8K & 209.8 & 148.6 & 203.7 & 1.0 \\
 & Tree Bench & 205.3 & 163.5 & 187.7 & -3.7 \\
 & OCR Bench & 163.6 & 204.4 & 218.2 & 2.7 \\
 & SEED 2 PLUS & 204.5 & 145.5 & 160.2 & 0.4 \\
 & CharXiv descriptive & 208.0 & 137.4 & 135.9 & 0.7 \\
 & CharXiv reasoning & 243.7 & 208.2 & 206.0 & -1.2 \\
 & Chart QA & 178.8 & 145.1 & 145.6 & 8.0 \\
 & MathVista & 277.1 & 278.9 & 265.2 & 0.2 \\
 & MathVerse & 404.1 & 362.0 & 359.6 & -4.8 \\
 & MathVision & 541.1 & 855.0 & 820.3 & -3.4 \\
 & WeMath & 371.8 & 380.3 & 384.7 & -5.2 \\
 & DynaMath & 374.1 & 269.1 & 273.4 & -0.4 \\
 & LogicVista & 380.6 & 298.7 & 300.9 & 4.5 \\
\bottomrule
\end{tabular}%
}
\end{table*}

\section{Use of AI Assistants}
We used ChatGPT for minor language polishing and readability improvements. The authors retained full responsibility for all scientific claims, analyses, and conclusions in the paper.







\end{document}